\documentclass[final,5p,times]{elsarticle}
\usepackage{adjustbox}
\usepackage{makecell}
\usepackage{amssymb}
\usepackage{amsmath}
\usepackage{lineno}
\usepackage{array}
\usepackage{float}
\usepackage[linesnumbered,ruled,vlined]{algorithm2e}
\usepackage{algorithmic}
\usepackage{subcaption}
\usepackage{tablefootnote}
\usepackage{url}
\usepackage{footmisc}
\usepackage[colorlinks = true, 
   linkcolor = blue,
   urlcolor = blue,
   citecolor = blue,
   anchorcolor = blue]{hyperref}

\usepackage{etoolbox}
\usepackage{pdflscape}
\usepackage{pifont}
\usepackage{tcolorbox}
\usepackage{tikz}
\usepackage{booktabs}
\usepackage{colortbl}
\usepackage{enumitem}
\usepackage{smartdiagram}
\usepackage{svg}

\AtBeginDocument{%
  \def\thefnote{\myfnsymbol{fnote}}}
\makeatletter
\def\myfnsymbol#1{\expandafter\@myfnsymbol\csname c@#1\endcsname}
\def\@myfnsymbol#1{\ifcase #1\or $\oplus$\or $\#\#$\else \@ctrerr\fi}
\def\fntext[#1]#2{\g@addto@macro\@fnotes{%
   \refstepcounter{fnote}\elsLabel{#1}%
   \def\thefootnote{\thefnote}
   \global\setcounter{footnote}{\c@fnote}%
   \footnotetext{#2}}}
\makeatother

\usepackage{ifthen}
\definecolor{arrowcolor}{RGB}{0, 0, 255}
\definecolor{textcolor}{RGB}{255, 0, 0}
\def\NTsize{\sffamily\large}
\def\CTsize{\sf\small}
\def\body#1#2#3#4#5#6#7#8#9{
        \begin{scope}[shift={(#1)},rotate=0,transform shape]
            \ifthenelse{\equal{#2}{end}}{
                \ifthenelse{\equal{#3}{reverse}}{
                    \filldraw[#6]
                        (-1pt,#5*0.5+#7/2)
                            -- ++(#4*0.5+1pt,0)
                            -- ++ (0,3mm) 
                            -- (#4,#5*0.5)
                            -- (#4*0.5,#5*0.5-#7/2-3mm)
                            -- ++(0,3mm)
                            -- ++(-#4*0.5-1pt,0);
                    }{
                    \filldraw[#6]
                        (#4+1pt,#5*0.5+#7/2)
                            -- ++(-#4*0.5-1pt,0)
                            -- ++ (0,3mm)
                            -- (0,#5*0.5)
                            -- (#4*0.5,#5*0.5-#7/2-3mm)
                            -- ++(0,3mm)
                            -- ++(#4*0.5+1pt,0);
                }
            }{
                \filldraw[#6] (0,#5*0.5+#7/2) rectangle ++(#4,-#7);
                \draw[textcolor](#4*0.5,#5*0.5) node[circle,draw,fill=white, minimum size=#7-3mm,font=\NTsize](temp){#2};
                \draw(temp)+(#8:#9) node[text width=#4*2, align=center,font=\CTsize](t2){#3};
                \draw[#6!40!white, preaction={draw,#6, line width=2pt}](temp)--(t2);
            }               
        \end{scope} 
}

\def\TAIL#1#2#3#4#5{
    \begin{scope}[shift={(#1)}]
        \filldraw[#4] (#2+1pt,#3*0.5+#5/2)-- ++(-#2*0.5-1pt,0) -- (#2*0.75,#3*0.5) -- (#2*0.5,#3*0.5-#5/2) -- ++ (1pt+#2*0.5,0);
    \end{scope} 
}

\def\CURVE#1#2#3#4#5#6{
    \begin{scope}[shift={(#1)},#4,transform shape]
        \filldraw[#5](-1pt,#3*0.5-#6/2) -- ++(1pt,0) arc (90:-90:#3*0.5-#6/2) -- ++(-1pt,0)-- ++(0,-#6) -- ++(1pt,0) arc (-90:90:#3/2+#6/2)--++(-1pt,0);
    \end{scope} 
}

\def\SNAKETEXT#1(#2)[#3][#4][#5]#6#7{
    \begin{scope}[shift={(#2)}]
        \edef\Shiftx{0}
        \edef\Shifty{0}
        \pgfmathparse{int(#6*2)}
        \xdef\tpl{\pgfmathresult}
        \foreach \cod/\direc/\dist/\desc [count=\ctr from 0] in {#1}{
            \ifnum\ctr=0
            \TAIL{\Shiftx*#3,-\Shifty*#4-#4}{#3}{#4}{#5}{#7}
            \fi
            \ifnum\ctr<#6
            \pgfmathparse{int(\Shiftx+1)}
            \xdef\Shiftx{\pgfmathresult}
            \body{\Shiftx*#3,-\Shifty*#4-#4}{\cod}{\desc}{#3}{#4}{#5}{#7}{\direc}{\dist}
            \fi
            \ifnum\ctr=#6
            \pgfmathparse{int(\Shiftx+1)}
            \xdef\Shiftx{\pgfmathresult}
            \CURVE{\Shiftx*#3,-\Shifty*#4-#4}{#3}{#4}{}{#5}{#7}
            \pgfmathparse{int(\Shiftx-1)}
            \xdef\Shiftx{\pgfmathresult}
            \pgfmathparse{int(\Shifty+1)}
            \xdef\Shifty{\pgfmathresult}
            \body{\Shiftx*#3,-\Shifty*#4-#4}{\cod}{\desc}{#3}{#4}{#5}{#7}{\direc}{\dist}
            \fi
            \ifnum\ctr>#6
            \ifnum\ctr<\tpl
            \pgfmathparse{int(\Shiftx-1)}
            \xdef\Shiftx{\pgfmathresult} 
            \body{\Shiftx*#3,-\Shifty*#4-#4}{\cod}{\desc}{#3}{#4}{#5}{#7}{\direc}{\dist}
            \fi
            \fi
            \ifnum\ctr=\tpl
            \pgfmathparse{int(\Shiftx-1)}
            \xdef\Shiftx{\pgfmathresult}
            \CURVE{\Shiftx*#3+#3,-\Shifty*#4-#4}{#3}{#4}{xscale=-1}{#5}{#7}
            \pgfmathparse{int(\Shifty+1)}
            \xdef\Shifty{\pgfmathresult}
            \pgfmathparse{int(\Shiftx+1)}
            \xdef\Shiftx{\pgfmathresult}
            \body{\Shiftx*#3,-\Shifty*#4-#4}{\cod}{\desc}{#3}{#4}{#5}{#7}{\direc}{\dist}
            \xdef\ctr{0}
            \fi
        }
    \end{scope} 
}

\usepackage{tocloft}

\usetikzlibrary{arrows.meta,shapes,positioning,shadows,trees}
\usepackage[edges]{forest}

\newcommand{\squaremarker}[1]{%
    \tikz[baseline=-0.75ex]\node[fill=#1, rectangle, minimum width=2.4mm, minimum height=2.4mm] {};%
}

\definecolor{SeafoodColor}{rgb}{1,0.56,0.5} 
\definecolor{FruitsColor}{rgb}{0.65,0.88,1} 
\definecolor{VegetablesColor}{rgb}{0.82,1,0.51} 

\begin{document}

\begin{frontmatter}

\title{Data clustering: a fundamental method in data science and management}

\author[a]{Tai Dinh\corref{cor1}\fnref{note}}
\author[a]{Wong Hauchi\fnref{note}}
\author[b]{Daniil Lisik}
\author [c] {Michal Koren}
\author[d]{Dat Tran}
\author[e]{Philip S. Yu}
\author[f]{Joaquín Torres-Sospedra}
\cortext[cor1]{Corresponding author: Tai Dinh (t\_dinh@kcg.ac.jp)}
\fntext[note]{Contributed equally to this work}

\affiliation[a]{organization={The Kyoto College of Graduate Studies for Informatics},
            addressline={7 Tanaka Monzencho, Sakyo Ward}, 
            city={Kyoto City},
            state={Kyoto},
            country={Japan}}
\affiliation[b]{organization={University of Gothenburg},
            addressline={Medicinaregatan 1F, 413 90}, 
            city={Göteborg},
            country={Sweden}}
\affiliation[c]{organization={Shenkar College of Engineering, Design and Art},
            addressline={Anne Frank St 12, Ramat Gan}, 
            city={Tel Aviv},
            country={Israel}}
\affiliation[d]{organization={University of Canberra},
            addressline={11 Kirinari St, Bruce ACT 2617}, 
            country={Australia}}
\affiliation[e]{organization={Department of Computer Science, University of Illinois at Chicago},
            addressline={Chicago},
            country={USA}}
\affiliation[f]{organization={Department of Computer Science,  University of Valencia},
            addressline={Valencia},
            country={Spain}}

\begin{abstract}
This paper explores the critical role of data clustering in data science, emphasizing its methodologies, tools, and diverse applications. Traditional techniques, such as partitional and hierarchical clustering, are analyzed alongside advanced approaches such as data stream, density-based, graph-based, and model-based clustering for handling complex structured datasets. The paper highlights key principles underpinning clustering, outlines widely used tools and frameworks, introduces the workflow of clustering in data science, discusses challenges in practical implementation, and examines various applications of clustering. By focusing on these foundations and applications, the discussion underscores clustering's transformative potential. The paper concludes with insights into future research directions, emphasizing clustering's role in driving innovation and enabling data-driven decision-making.
\end{abstract}

 
\end{frontmatter}



\section{Introduction} \label{sec:introduction}
\begin{figure*}[!htb]
\vspace{-2.5cm}
  \centering
  \includegraphics[width=0.8\linewidth]{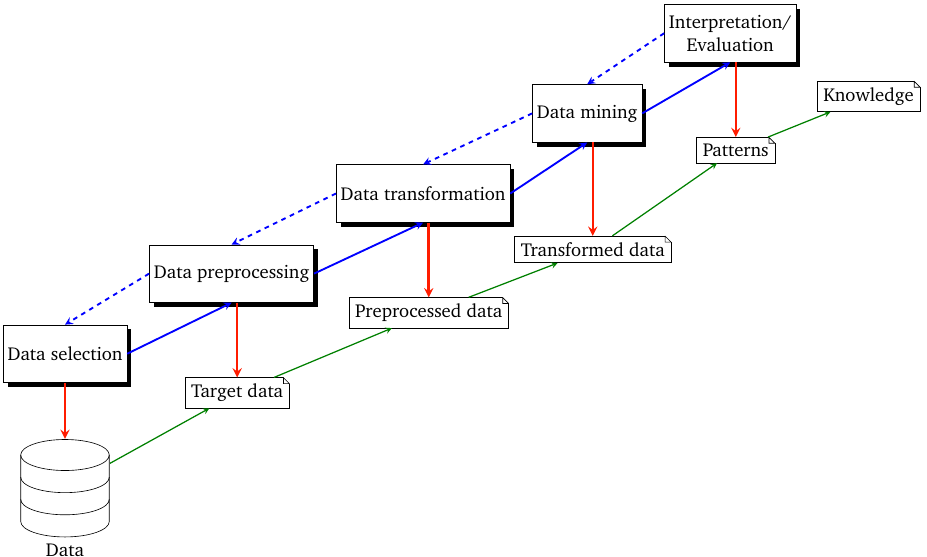}
  \caption{A general structure of \textsc{Knowledge Discovery in Databases}}
  \label{fig:kdd_process}
\end{figure*}

Advances in information and data acquisition technologies allow the daily collection of vast amounts of data from sources like sensors, the internet, and devices. These data reflect system behaviors and may contain valuable knowledge. \textsc{Knowledge Discovery in Databases (KDD)} uses \textsc{Data Mining} and related methods to find useful patterns in large datasets. Data mining, also called knowledge extraction, applies specific algorithms for pattern detection. Figure \ref{fig:kdd_process} illustrates the general KDD process as described in \citep{abonyi2007cluster}, highlighting the role of \textsc{Data Mining} as a key step in the overall process of knowledge discovery from data.

\textsc{Data Science} is a broader, interdisciplinary field that combines computer science, statistics, mathematics, domain expertise, and other techniques to extract insights and derive solutions from data \citep{kotu2018data}. While \textsc{Data Mining} focuses on discovering patterns and relationships within datasets, it is just one step within the larger \textsc{Data Science} workflow. In the context of KDD, \textsc{Data Science} extends beyond pattern detection to include data preprocessing, visualization, predictive modeling, and decision-making, offering a comprehensive framework for solving complex problems with data.

\textsc{Clustering} is a fundamental technique in \textsc{Data Science}, which organizes data into meaningful groups, or clusters, based on their intrinsic similarities \citep{aggarwal2013introduction}. A cluster is generally considered as a group of objects in which objects within each cluster are more closely related to one another than objects assigned to other different clusters.
Unlike classification, which relies on predefined labels, clustering is an unsupervised learning approach. It is ``unsupervised" because it is not guided by a priori ideas of which variables or samples belong in which clusters, and ``learning" because the machine algorithms learn how to cluster \citep{kassambara2017practical}. This technique is widely used to handle complex, high-dimensional datasets and is pivotal for understanding the inherent organization of data \citep{han2022data}.

\begin{figure}[!htb]
  \centering
  \includegraphics[width=\linewidth]{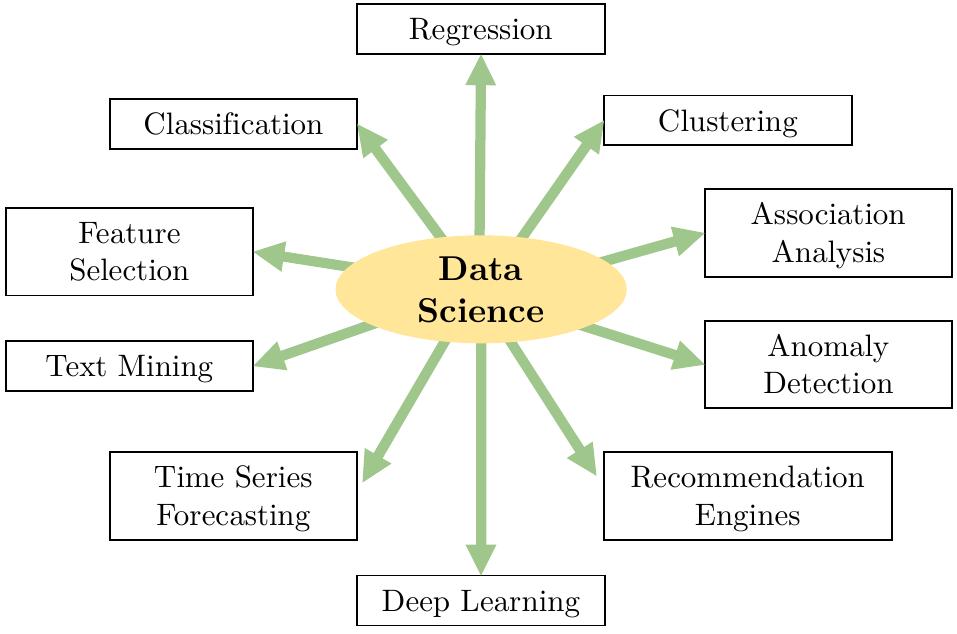}
  \caption{Data science tasks \citep{kotu2018data}}
  \label{fig:data_science}
\end{figure}

In today’s data-driven world, where organizations and researchers deal with enormous volumes of data, clustering serves as a critical tool for summarizing and analyzing this information. By grouping similar data points, clustering aids in simplifying complex datasets, enabling better exploration and visualization, and laying the groundwork for further data analysis \citep{tukey1977exploratory}. Furthermore, clustering finds applications in a wide array of fields. The objects to be clustered may include words, text, images, audio, video, longitudinal media data, graph nodes, or any collection of items that can be described by a set of features \citep{izenman2008}. As a versatile tool, clustering has been widely employed as an intermediate step in numerous data mining and machine learning tasks. As illustrated in Figures \ref{fig:data_science}, clustering is recognized as a fundamental branch of machine learning and a pivotal task in the broader field of data science.

Clustering has been applied across diverse domains, including natural sciences, social sciences, economics, education, engineering, and health science, to name a few \citep{dinh2024categorical}. Additionally, clustering is closely connected to other data science processes, such as regression and classification. For instance, clustering can enhance regression analysis by identifying distinct groups within data, which may then inform separate regression models for each cluster. Similarly, clustering can complement classification by identifying potential class labels or refining existing ones based on patterns in the data. By serving as both a foundational and complementary tool, clustering facilitates a deeper understanding of data, enabling more accurate predictions, targeted interventions, and efficient resource allocation.

This paper provides a comprehensive exploration of data clustering within the context of data science. It begins by delving into the fundamental concepts that underpin clustering, including key definitions and principles. The paper then examines various clustering methodologies, ranging from traditional techniques such as partitional and hierarchical clustering to more advanced approaches like data stream and subspace clustering. Finally, the discussion extends to the diverse applications of clustering, highlighting its impact across various fields and its potential to foster innovation. By presenting a detailed overview of these aspects, this paper aims to highlight the importance of data clustering in data science and offer readers a comprehensive understanding of this essential technique.

The remainder of this paper is organized as follows: Section \ref{sec:basic_data_clustering} provides an overview of clustering fundamentals, including a taxonomy of clustering methodologies and validation metrics. Section \ref{sec:algorithms} introduces commonly used clustering algorithms, frameworks, libraries, tools, and the clustering workflow. Section \ref{sec:challenges} addresses key challenges in practical implementation. Section \ref{sec:applications} explores various applications of clustering. Finally, Section \ref{sec:conclusion} summarizes the paper and outlines potential directions for future work.

\section{Basic of data clustering} \label{sec:basic_data_clustering}
\subsection{A taxonomy of clustering methodologies}
Clustering is an unsupervised learning process that partitions a dataset \(\displaystyle D = \{x_1, x_2, \ldots, x_n\}\) where each data point \(\displaystyle x_i \in \mathbb{R}^d\), into \(k\) clusters \(\displaystyle C = \{C_1, C_2, \ldots, C_k\}\). The goal is to group similar data points into the same cluster while ensuring that points in different clusters are dissimilar. Formally, the clustering task can be expressed as the optimization of a clustering objective \(\displaystyle \mathcal{F}(C)\), 
where:

\[
\begin{cases}
\displaystyle \bigcup_{j=1}^k C_j = D, 
\quad C_i \cap C_j = \varnothing \quad \text{for} \quad i \neq j 
\quad (1) \\[1em]
\displaystyle \arg\min_{C} \, \mathcal{F}(C) 
\quad (2)
\end{cases}
\]

Equation (1) indicates that clusters are non-overlapping and exhaustive, whereas equation (2) aims to enforce that each cluster \(\displaystyle C_j\) is defined such that points within the same cluster are more similar to each other (intra-cluster similarity) than those in different clusters (inter-cluster dissimilarity).

\begin{figure*}[!htb]
\vspace{-2cm}
  \centering
  \includegraphics[width=0.8\linewidth]{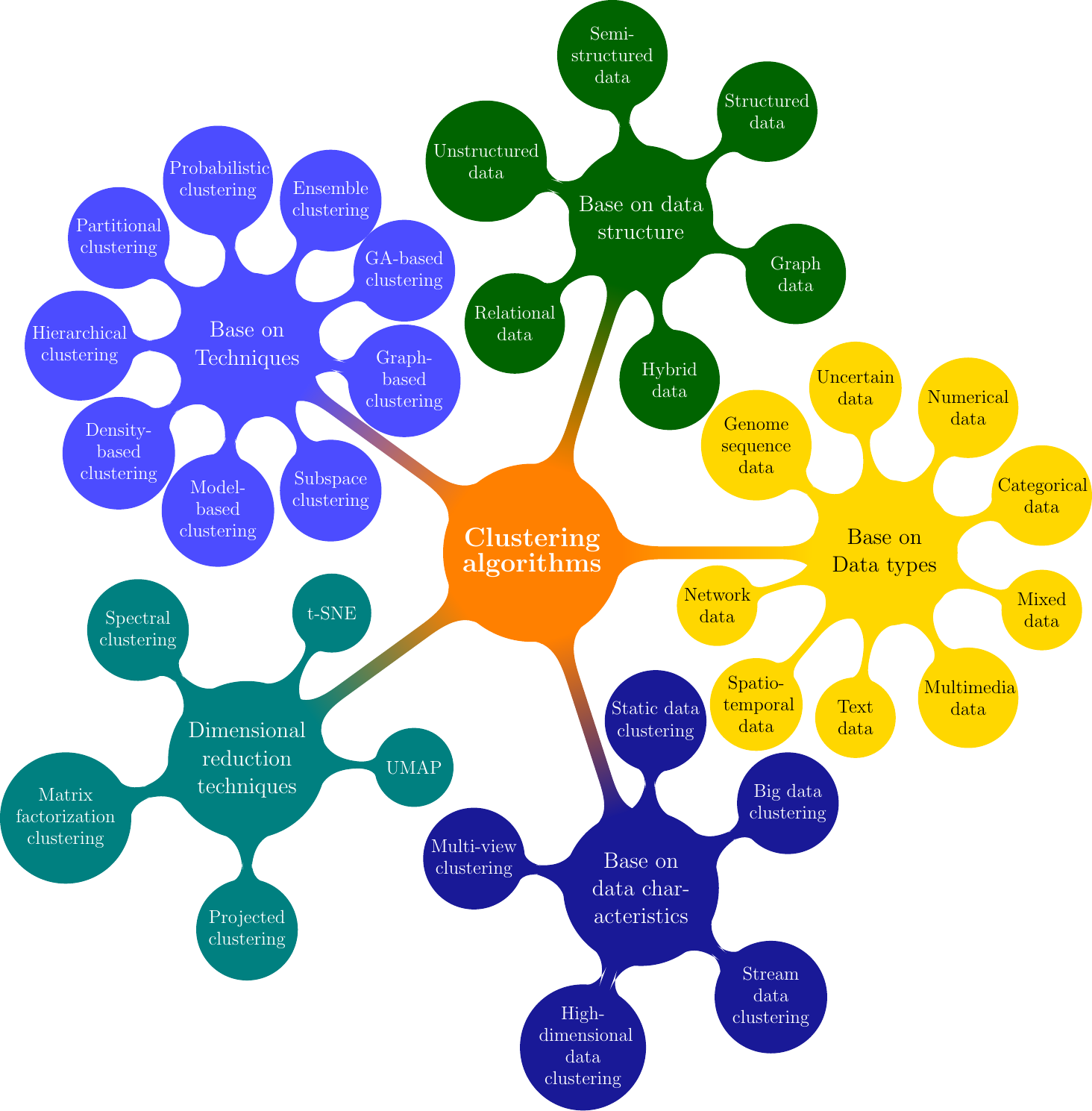}
  \caption{A classification of clustering algorithms}
  \label{fig:clustering_classification}
\end{figure*}

Figure \ref{fig:clustering_classification} shows a taxonomy of clustering algorithms. In what follows, we revisit some of the most commonly used types of clustering discussed in the literature.

\begin{figure}[!htb]
  \centering
  \includegraphics[width=\linewidth]{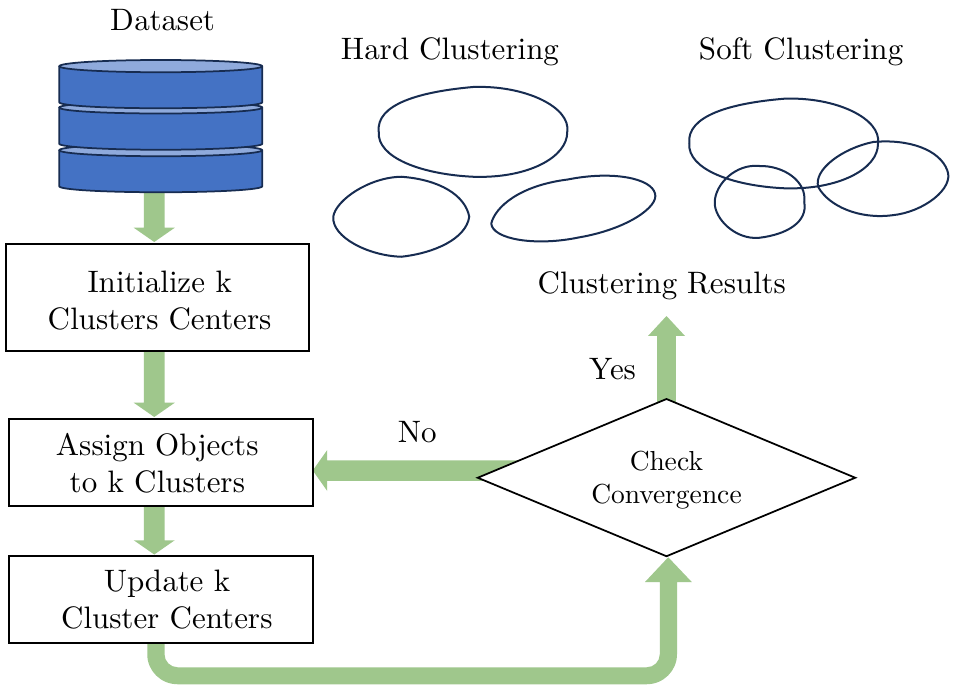}
  \caption{\textsc{Partitional Clustering}}
  \label{fig:partitional_clustering}
\end{figure}

\begin{figure}[!htb]
  \centering
  \includegraphics[width=\linewidth]{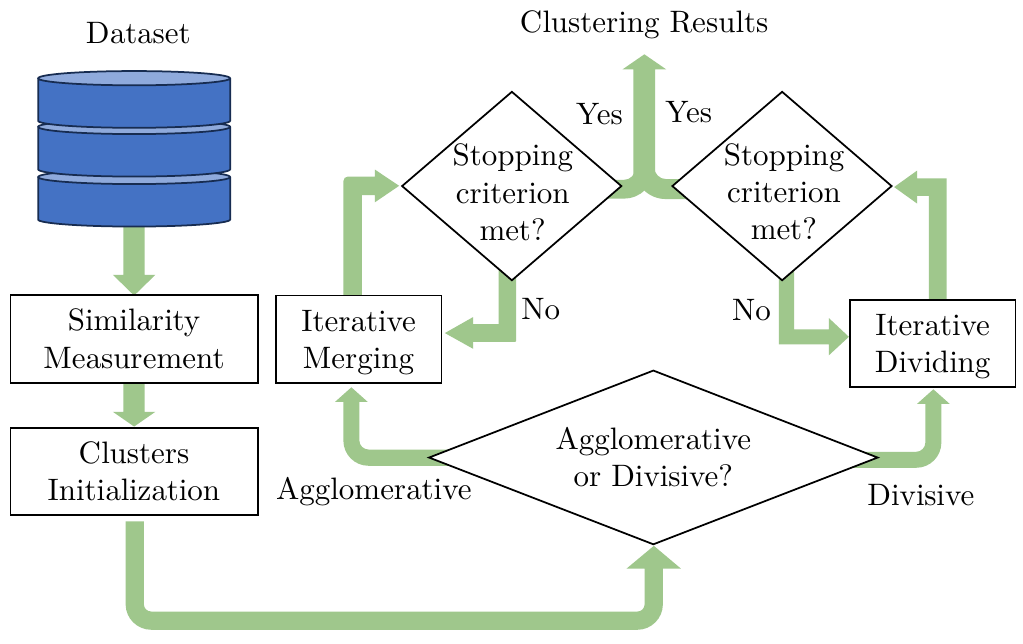}
  \caption{\textsc{Hierarchical Clustering}}
  \label{fig:hierarchical_clustering}
\end{figure}

Partitional (partitioning) clustering, also known as nonhierarchical or flat clustering, partitions a dataset into a predefined number of clusters k, without constructing hierarchical structures. Figure \ref{fig:partitional_clustering} shows the general workflow of partitional clustering algorithms. The primary goal is to optimize an objective function, such as minimizing intra-cluster distances or maximizing inter-cluster separation \citep{kaufman1990finding}. 
Partitional clustering algorithms can be broadly categorized into hard partitional clustering and soft partitional clustering. Hard clustering, also referred to as crisp clustering, assigns each data point to exactly one non-overlapping cluster. Algorithms like \textsc{K-Means} \citep{lloyd1982least}, \textsc{K-Modes} and \textsc{K-Prototypes} \citep{huang1998extensions} fall under this category. In contrast, soft (or fuzzy) clustering allows data points to belong to multiple clusters simultaneously. Each point is assigned a membership value ranging from 0 (no membership) to 1 (full membership), indicating its degree of association with each cluster \citep{tan2019introduction}. This flexibility is particularly useful for datasets with overlapping or ambiguous boundaries.

Hierarchical clustering constructs a dendrogram to represent relationships among data points, using either an agglomerative (bottom-up) or divisive (top-down) approach \citep{han2022data}. Figure \ref{fig:hierarchical_clustering} shows the workflow of hierarchical clustering algorithms. In agglomerative clustering, each object initially starts as a separate cluster, and the two closest clusters are iteratively merged based on a similarity measure until only one cluster remains. Conversely, divisive clustering starts with all objects in a single cluster and recursively splits them into smaller clusters. The dendrogram illustrates the merging or splitting process, and the optimal number of clusters can be determined by cutting the dendrogram at a level that balances the number of clusters and their homogeneity.

\begin{figure}[!htb]
  \centering
  \includegraphics[width=\linewidth]{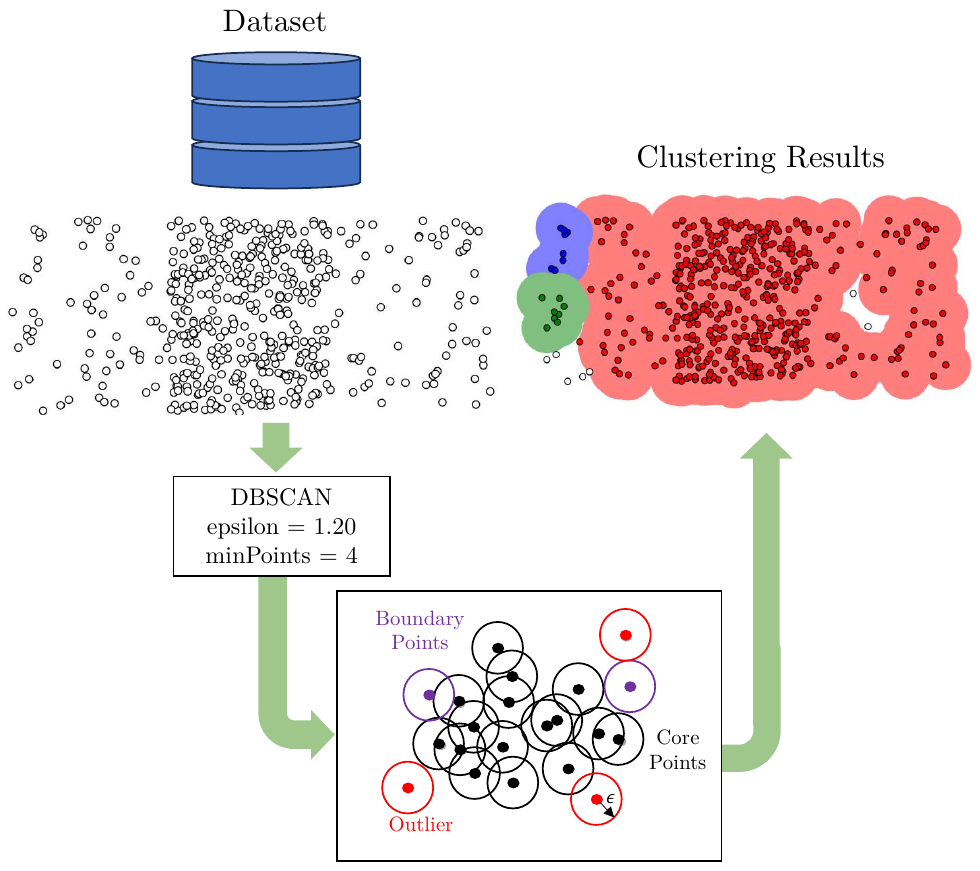}
  \caption{\textsc{Density-based Clustering}}
  \label{fig:density_clustering}
\end{figure}

Density-based clustering identifies clusters as dense regions of data points separated by sparser areas. Figure \ref{fig:density_clustering} shows the workflow of density-based clustering. Unlike partitional or hierarchical methods, it does not require specifying the number of clusters in advance. Instead, clusters are formed based on the density of points within a defined neighborhood. This approach is particularly effective for discovering clusters of arbitrary shapes and handling noise in datasets. A prominent example is the \textsc{DBSCAN} algorithm \citep{ester1996density}, which groups points with sufficient density while labeling outliers as noise.

\begin{figure}[!htb]
  \centering
  \includegraphics[width=0.9\linewidth]{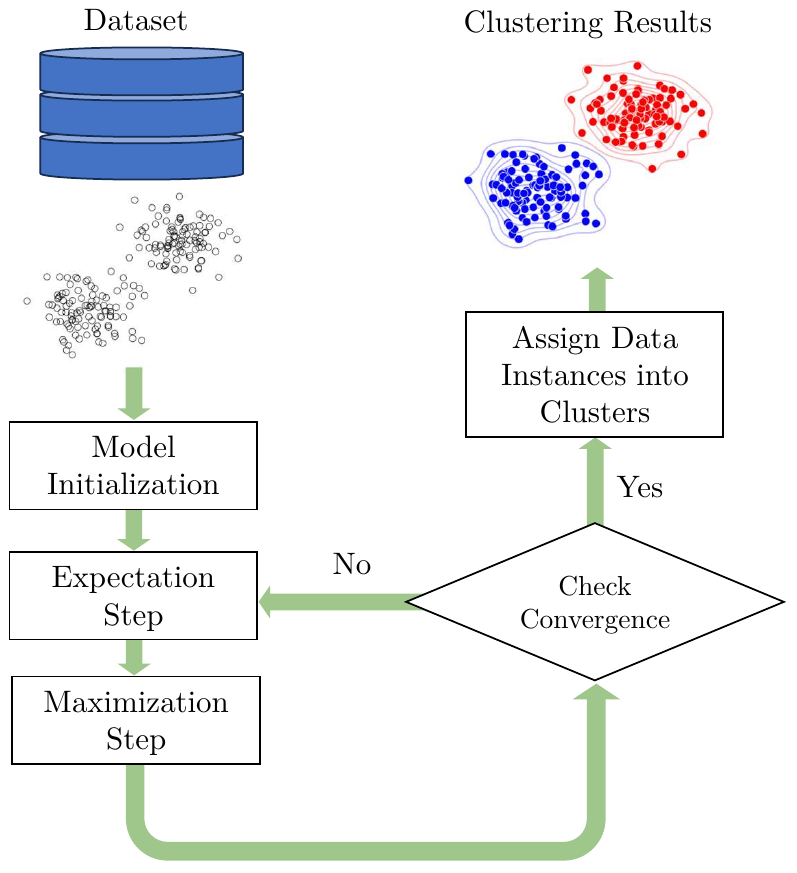}
  \caption{\textsc{Model-based Clustering}}
  \label{fig:model_clustering}
\end{figure}

Model-based clustering is a statistical approach that assumes data is generated from a mixture of underlying probability distributions, with each distribution corresponding to a specific cluster \citep{fraley2002model}. Figure \ref{fig:model_clustering} illustrates the workflow of a model-based clustering algorithm. This approach relies on a generative model in which clusters are represented by parametric probability distributions, such as Gaussian distributions in the widely used \textsc{Gaussian Mixture Model} (GMM). The method requires specifying distributional assumptions, as these directly influence the clustering outcome by defining the shape, size, and overlap of clusters. The clustering process involves estimating the parameters of these distributions and assigning data points to the cluster that maximizes their likelihood. By leveraging statistical principles, model-based clustering can effectively capture complex data structures, determine the optimal number of clusters using model selection criteria (e.g., Bayesian Information Criterion (BIC) or Akaike Information Criterion (AIC)), and handle overlapping clusters. However, the validity of the results depends on whether the chosen distributional assumptions align with the true underlying data characteristics.

\begin{figure}[!htb]
  \centering
  \includegraphics[width=\linewidth]{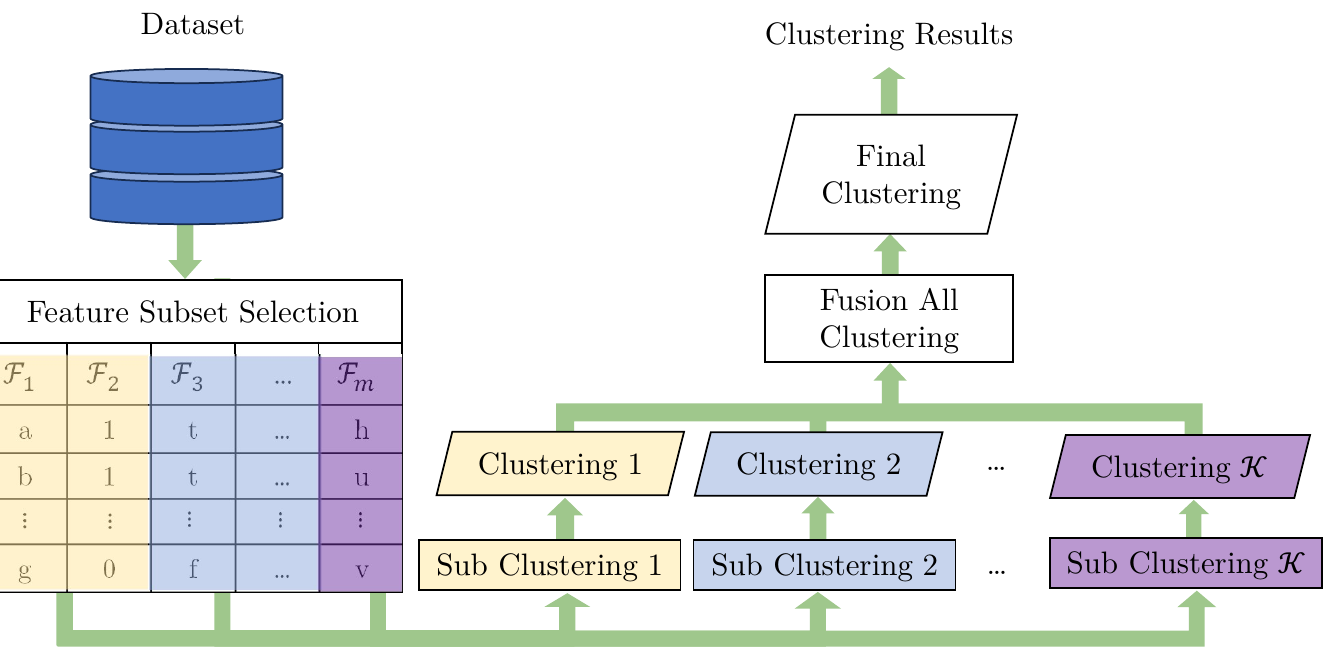}
  \caption{\textsc{Subspace Clustering}}
  \label{fig:subspace_clustering}
\end{figure}

Subspace clustering focuses on identifying clusters within specific subspaces of a high-dimensional dataset, rather than the full-dimensional space \cite{parsons2004subspace}. Figure \ref{fig:subspace_clustering} shows the workflow of subspace clustering. The key idea is that meaningful clusters may exist in subsets of dimensions where data points share strong similarities, while other dimensions may introduce noise. By searching for clusters in these relevant subspaces, this approach is particularly effective for high dimensional data where traditional methods often struggle. Subspace clustering has proven valuable in domains such as bioinformatics and text analysis, where data often exhibit localized patterns in specific subsets of features.

\begin{figure}[!htb]
  \centering
  \includegraphics[width=\linewidth]{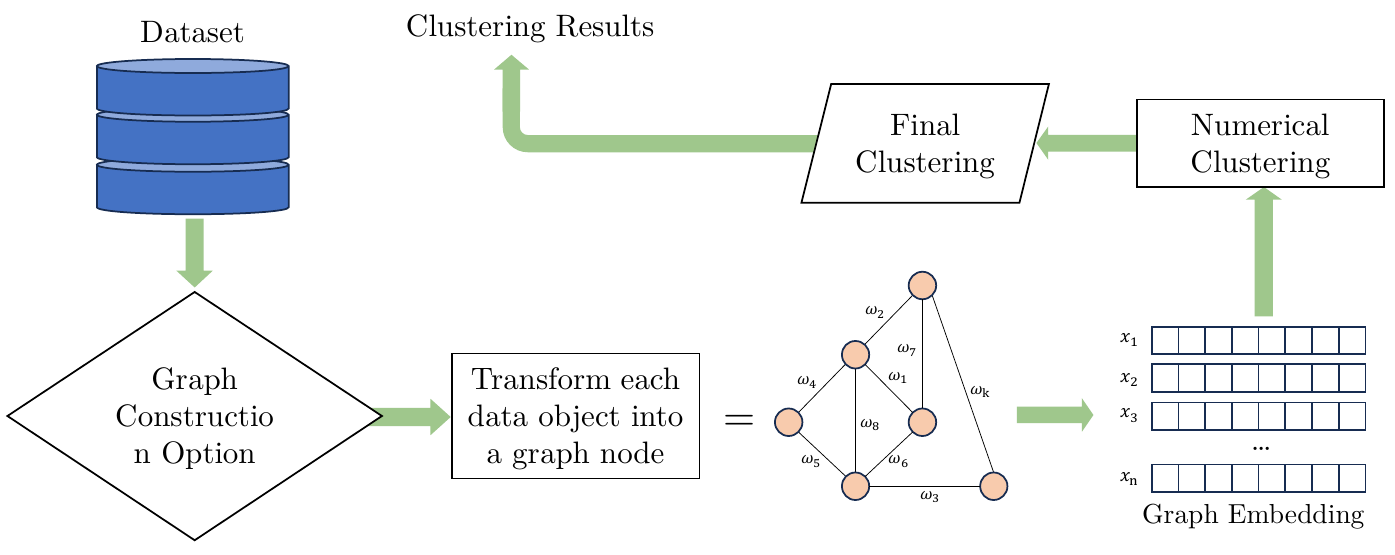}
  \caption{\textsc{Graph Clustering}}
  \label{fig:graph_clustering}
\end{figure}

Graph-based clustering represents data as a graph, where nodes correspond to data points and weighted edges reflect the similarity between them \citep{maier2008influence}. Figure \ref{fig:graph_clustering} shows the workflow of graph-based clustering. Clusters are formed as densely connected subgraphs, with sparse or minimal connections to nodes outside the cluster. The goal of graph-based clustering algorithms is to partition the graph by analyzing its edge structure, typically maximizing the total weight or number of edges within clusters while minimizing inter-cluster connections. This approach is particularly effective for identifying non-convex clusters and handling complex data structures.

\begin{figure}[!htb]
  \centering
  \includegraphics[width=\linewidth]{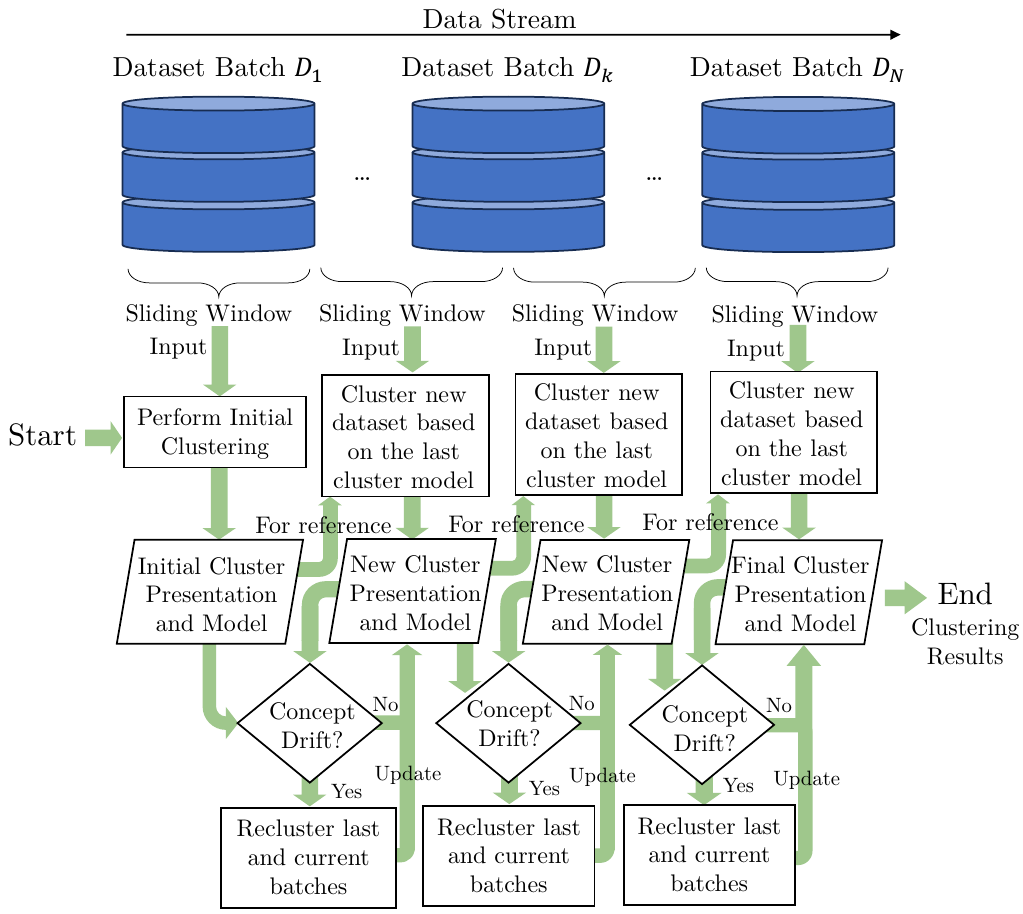}
  \caption{\textsc{Data Stream Clustering}}
  \label{fig:stream_clustering}
\end{figure}

Data stream clustering dynamically groups similar data points in real-time as they flow continuously into a system \cite{silva2013data}. Designed to handle high-velocity, large-volume, and unbounded data streams, it incrementally updates clusters to ensure efficient and adaptive processing. The general framework of data stream clustering, illustrated in Figure \ref{fig:stream_clustering}, comprises several key steps. It begins with clustering the initial batch of data to establish a baseline configuration. A sliding window model is then employed to focus on the most recent data, ensuring that older, less relevant points are discarded. Clusters are represented through statistical summaries, such as centers or spreads, which are continuously updated to reflect new data. As new points arrive, they are incrementally assigned to existing clusters or used to form new ones, depending on their similarity. The framework also incorporates mechanisms to detect and respond to concept drift, adjusting the clustering model to accommodate shifts in the data distribution. Periodic maintenance is performed to refine clusters, which may involve merging, splitting, or repositioning them as needed. This iterative process ensures that the clustering remains adaptive, providing an up-to-date representation of the evolving data stream for downstream analysis.

\begin{figure}[!htb]
  \centering
  \includegraphics[width=0.9\linewidth]{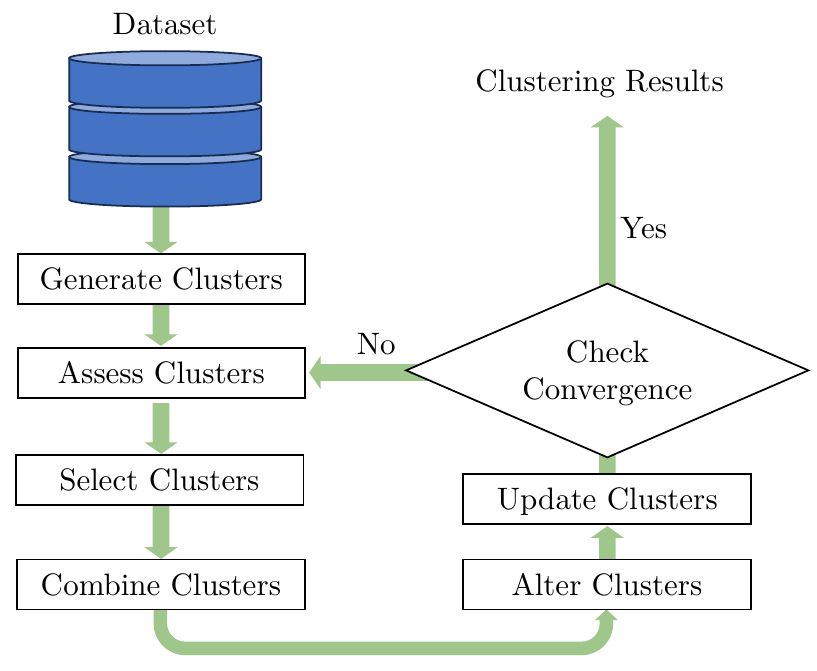}
  \caption{\textsc{Genetic Algorithm-based Clustering}}
  \label{fig:ga_clustering}
\end{figure}

\begin{figure}[!htb]
  \centering
  \includegraphics[width=\linewidth]{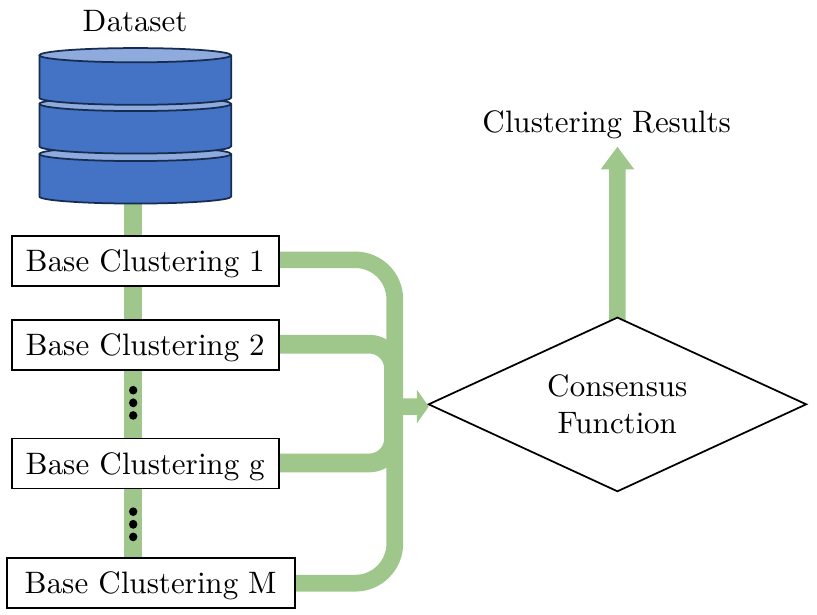}
  \caption{\textsc{Ensemble Clustering}}
  \label{fig:ensemble_clustering}
\end{figure}

\begin{figure*}[!htb]
\vspace{-2.5cm}
  \centering
  \includegraphics[width=0.9\linewidth]{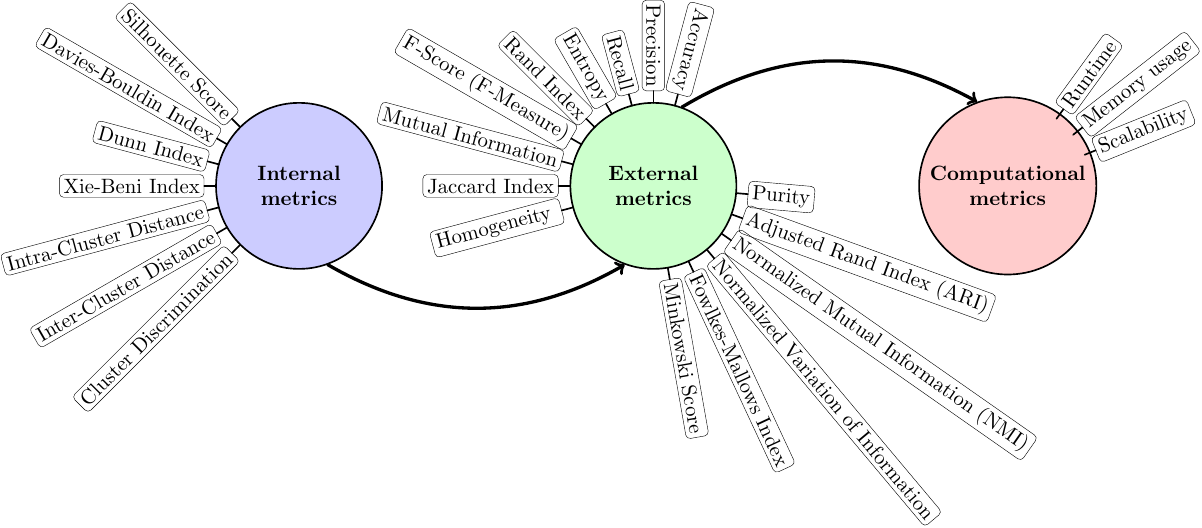}
  \caption{A taxonomy of common evaluation metrics for clustering}
  \label{fig:metrics_classification}
\end{figure*}

Genetic algorithm (GA)-based clustering \citep{maulik2000genetic} utilizes genetic algorithms (GAs) to improve clustering performance. GAs are search heuristics inspired by the process of natural selection, operating on a population of potential solutions and evolving them through selection, crossover, and mutation \citep{holland1992adaptation}. Figure \ref{fig:ga_clustering} shows the framework of GA-based clustering. In clustering, each potential solution (or chromosome) represents a partitioning of the dataset into clusters, where each cluster contains a subset of data points. By iteratively optimizing cluster assignments through evolutionary processes, GA-based methods offer robust solutions and reduce the risk of being trapped in local optima.

Ensemble clustering combines multiple clustering outputs---either from different algorithms, multiple independent executions of the same algorithm (e.g., with varying initializations), or runs on subsets of the data---to produce a single consensus partition of the original dataset \citep{strehl2002cluster}. The workflow of ensemble clustering is illustrated in Figure \ref{fig:ensemble_clustering}. A typical ensemble framework consists of two main steps: (1) generating a set of base clustering results and (2) integrating these results into a final consensus clustering using a consensus function. Both the quality of the base clusterings and the effectiveness of the consensus function play a critical role in determining the success of the ensemble. Additionally, a sensible combination of algorithms should be chosen, as different clustering algorithms operate under varying assumptions about the data and have distinct goals or definitions of optimal performance.
\subsection{Clustering validation metrics} \label{sec:evaluation}

Figure \ref{fig:metrics_classification} illustrates common metrics for assessing the performance of clustering algorithms. Internal validation metrics assess the quality of clustering algorithms by evaluating their performance based solely on the data used for clustering, without external references. External validation metrics, on the other hand, compare clustering results to an external ground truth or reference standard. In addition to quality-based metrics, the computational performance of clustering algorithms is also a critical consideration. \textsc{Runtime} measures the time required for an algorithm to process the input dataset and produce outputs. A lower runtime indicates better computational efficiency. \textsc{Scalability} assesses how the performance of clustering algorithms changes with varying dataset sizes. Good scalability implies that an algorithm maintains consistent performance across different data volumes. \textsc{Memory usage} quantifies the amount of memory required to execute a specific algorithm. Lower memory usage indicates better resource efficiency. While lower runtime and memory usage generally indicate a more efficient clustering algorithm in terms of computational complexity, these metrics should be balanced against the quality of clustering results, as assessed by internal and external validation metrics.
\section{Data clustering algorithms}\label{sec:algorithms}
\subsection{Popular clustering algorithms}
\begin{figure}[!htb]
  \centering
  \includegraphics[width=0.8\linewidth]{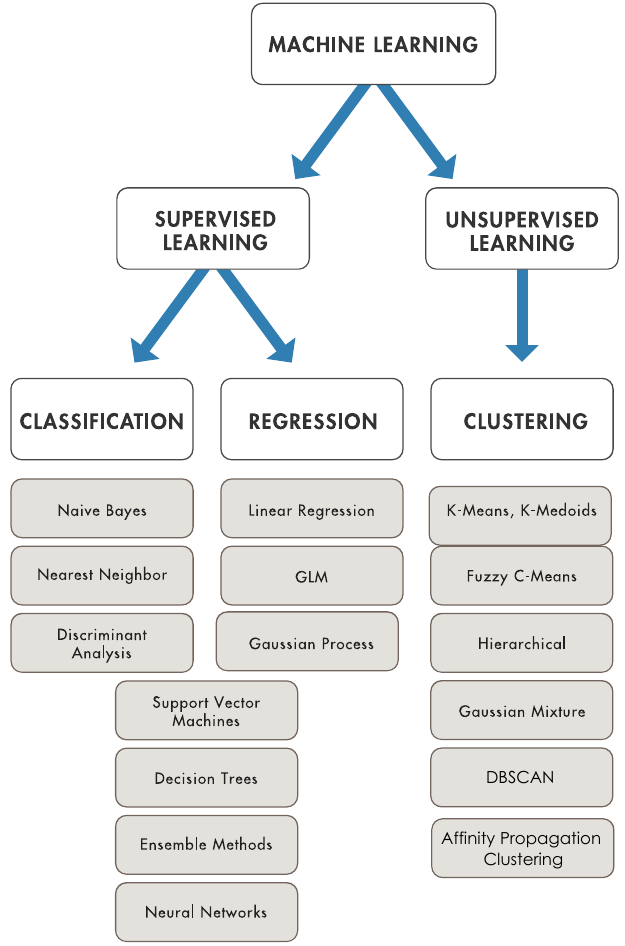}
  \caption{A taxonomy of machine learning algorithms \citep{matlab}}
  \label{fig:machine_learning}
\end{figure}
\begin{figure*}[!htb]
\vspace{-2cm}
    \centering
    \begin{adjustbox}{max width=\linewidth}
    \begin{tikzpicture}
        \SNAKETEXT{
            1957/90/15mm/ \textsc{K-Means} \citep{lloyd1982least} \squaremarker{blue},
            1963/-90/15mm/ \textsc{Ward Linkage HAC} \citep{ward1963hierarchical} \squaremarker{pink},
            1967/90/15mm/ \textsc{K-Means} \citep{macqueen1967some} \squaremarker{blue},
            1973/-90/15mm/ \textsc{Single Linkage HAC} \citep{sibson1973slink} \squaremarker{pink},
            1977/90/15mm/ \textsc{Complete Linkage HAC} \citep{defays1977efficient} \squaremarker{pink},
            1979/-90/15mm/ \textsc{K-Means} \citep{hartigan1979k} \squaremarker{blue},
            1982/90/15mm/ \textsc{SOMs} \citep{kohonen1982self}  \squaremarker{yellow},
            1984/-90/15mm/ \textsc{Fuzzy C-Means} \citep{bezdek1984fcm}  \squaremarker{green},
            1990/90/15mm/ \textsc{PAM} \citep{kaufman1990finding}  \squaremarker{blue},
            1995/-90/15mm/ \textsc{Mean Shift} \citep{cheng1995mean}  \squaremarker{red},
            1996/90/15mm/ \textsc{DBSCAN} \citep{ester1996density}  \squaremarker{red},
            1996/-90/15mm/ \textsc{BIRCH} \citep{zhang1996birch}  \squaremarker{pink},
            1997/90/15mm/ \textsc{K-Modes} \citep{huang1997fast}  \squaremarker{blue},
            1998/-90/15mm/\textsc{K-Prototypes} \citep{huang1998extensions}  \squaremarker{blue},
            1998/90/15mm/\textsc{CLIQUE} \citep{agrawal1998automatic} \squaremarker{orange},
            1998/-90/15mm/ \textsc{MClust} \citep{fraley1998mclust} \squaremarker{yellow},
            1999/90/15mm/ \textsc{Fuzzy K-Modes} \citep{huang1999fuzzy}  \squaremarker{green},
            1999/-90/15mm/ \textsc{Genetic K-Means} \citep{krishna1999genetic}  \squaremarker{purple},
            1999/90/15mm/ \textsc{OPTICS} \citep{ankerst1999optics}  \squaremarker{red},
            2000/-90/15mm/ \textsc{Rock} \citep{guha2000rock}  \squaremarker{pink},
            2001/90/15mm/ \textsc{Spectral Clustering} \citep{ng2001spectral}  \squaremarker{brown},
            2002/-90/15mm/ \textsc{Cluster Ensemble} \citep{strehl2002cluster}  \squaremarker{cyan},
            2003/90/15mm/ \textsc{LDA} \citep{blei2003latent} \squaremarker{yellow},
            2003/-90/15mm/ \textsc{CluStream} \citep{aggarwal2003framework}  \squaremarker{gray},
            2007/90/15mm/ \textsc{Affinity Propagation} \citep{frey2007clustering}  \squaremarker{brown},
            2008/-90/15mm/ \textsc{Louvain} \citep{blondel2008fast}  \squaremarker{brown},
            2008/90/15mm/ \textsc{T-SNE} \citep{van2008visualizing}  \squaremarker{black},
            2017/-90/15mm/ \textsc{HDBSCAN} \citep{mcinnes2017hdbscan}  \squaremarker{red},
            2018/90/15mm/ \textsc{UMAP} \citep{mcinnes2018umap}  \squaremarker{black},
            end/reverse%
        }(0,0)[30mm][47mm][arrowcolor]{6}{10mm}
    \end{tikzpicture}
    \end{adjustbox}
    \begin{tabular}{ccccc}
    \\
    \squaremarker{blue} Hard partitional & \hspace{0.3cm} \squaremarker{green} Fuzzy partitional & \squaremarker{pink} Hierarchical & \hspace{0.2cm}\squaremarker{red} Density-based & \hspace{0.2cm} \squaremarker{gray} Data stream  \hspace{0.2cm} \squaremarker{purple} Genetic Algorithm (GA)-based \\
    \hspace{-0.9cm}\squaremarker{orange} Subspace &  \hspace{-0.7cm} \squaremarker{cyan} Ensemble &  \hspace{0.1cm}\squaremarker{brown} Graph-based &  \squaremarker{yellow} Model-based & \hspace{-3cm} \squaremarker{black} Dimensionality reduction\\
    \end{tabular}
    \caption{Timeline of common data clustering algorithms}
    \label{fig:historical_development}
\end{figure*}
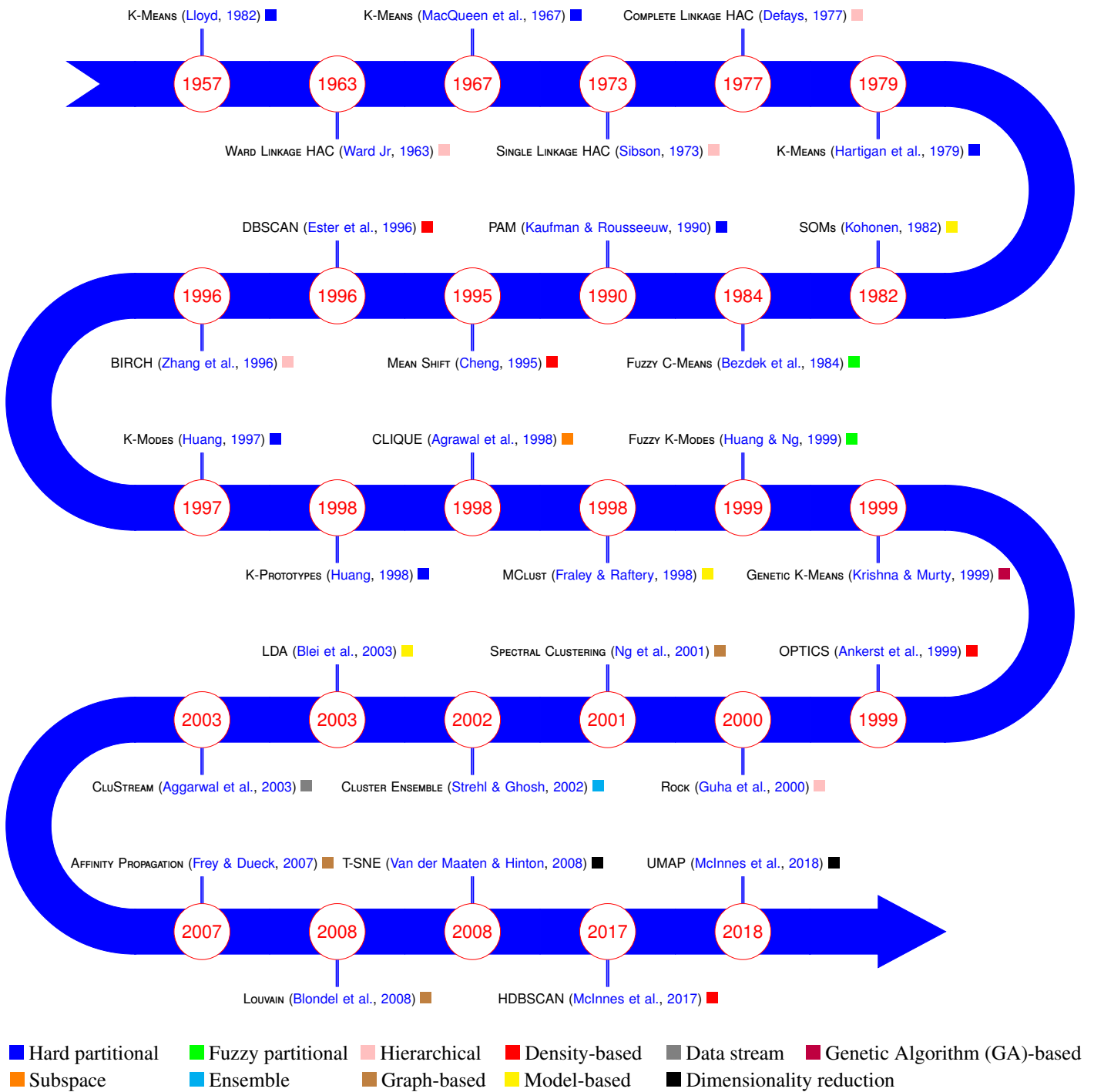

In machine learning, many clustering algorithms as shown in Figure \ref{fig:machine_learning} have been developed to approach existing problems users face from different perspectives. These algorithms vary in their methodologies, assumptions, and the data types they handle most effectively. Some focus on minimizing distances between data points, while others prioritize density or hierarchical relationships. As data science continues to evolve, more and more algorithms have become integral tools in fields ranging from market segmentation to anomaly detection. 

Figure \ref{fig:historical_development} illustrates the progression of clustering algorithms over time. The figure highlights algorithms that are widely recognized and commonly utilized within the research community. In the subsequent sections, we will introduce some of the most prominent clustering algorithms, emphasizing their key features and applications.

\textsc{K-Means} \citep{lloyd1982least} is a popular algorithm of partitioning methods used for clustering data into groups based on similarity. Lloyd's algorithm (Algorithm \ref{algo:lloyd}) is the most common implementation of \textsc{K-Means}. It begins by initializing $k$ cluster centroids, which are typically chosen randomly. Each data point is then assigned to the nearest centroid based on a distance metric, usually \textsc{Euclidean\ distance}, forming $k$ clusters. Once the assignments are made, the centroids are updated by calculating the mean position of all points in each cluster. These steps of assignment and update are repeated iteratively until the centroids stabilize or a stopping criterion is met, such as a maximum number of iterations or minimal movement of centroids.

\begin{algorithm}
\DontPrintSemicolon
\SetAlgoLined
\caption{Lloyd's \textsc{K-Means} \citep{lloyd1982least}}
    Initialize by randomly assigning each data point to one of the \(k\) clusters\;
    \Repeat{no data points change clusters (convergence) or centroids no longer change}{
        Calculate the centroids of each cluster as the mean of all data points currently assigned to that cluster\;
        Reassign each data point to the cluster with the nearest centroid\;
    }
\label{algo:lloyd}
\end{algorithm}

\cite{macqueen1967some} provides another variant of \textsc{K-Means} as shown in Algorithm \ref{algo:macqueen}. Instead of recalculating the centroids after all points have been assigned (as in Lloyd’s algorithm), MacQueen's algorithm updates centroids incrementally as each point is assigned to a cluster. This method may converge faster in some cases due to incremental updates. However, it is not always faster than Lloyd’s, and it might be more sensitive to the order of data points, especially in larger datasets \citep{izenman2008,reddy2013survey}.

\begin{algorithm}
\DontPrintSemicolon
\SetAlgoLined
\caption{MacQueen's \textsc{K-Means} \citep{macqueen1967some}}
    Randomly select $k$ data points as initial centroids\;
    \Repeat{convergence}{
        \For{each data point $x_i$}{
            Assign $x_i$ to the nearest centroid $c_j$\;
            Update the centroid $c_j$ as mean over all data points assigned to it so far, including $x_i$\;
        }
    }
\label{algo:macqueen}
\end{algorithm}

Generally, \textsc{K-Means} is efficient with large datasets and is computationally straightforward, with a time complexity that generally scales linearly with the number of data points \citep{aggarwal2013introduction}. However, it often converges to a local rather than a global optimum, depending on the initial centroid positions \citep{macqueen1967some}. In addition, the algorithm tends to form clusters with convex, spherical shapes, making it less suitable for complex or non-convex structures \citep{anderberg1973cluster}. Despite these limitations, \textsc{K-Means} remains popular due to its simplicity and effectiveness in many practical applications, such as customer segmentation, document clustering, pattern recognition, and data preprocessing, to name a few \citep{han2022data,wu2008top}.

\begin{algorithm}
\caption{\textsc{K-Modes} \citep{huang1998extensions}}
\DontPrintSemicolon
\SetAlgoLined
    Initialize by selecting $k$ initial \emph{modes}, one for each cluster\;
    Allocate each data object to the cluster whose \emph{mode} is the nearest according to the \textsc{Simple Matching Dissimilarity} measure\;
    Update the \emph{mode} of the cluster after each allocation\;
    \Repeat{convergence}{
        \For{each object in the dataset}{
            Retest the dissimilarity of the object against the current \emph{modes}\;
            \If{the nearest \emph{mode} for the object belongs to a different cluster}
            {
                Reallocate the object to that cluster\;
                Update the \emph{modes} of both affected clusters\;
            }
        }
    }
\label{algo:kmodes}
\end{algorithm}

\textsc{K-Modes} \citep{huang1998extensions} is another partitioning methods of clustering technique, extending the \textsc{K-Means}  algorithm. Instead of using the mean to define cluster centroids, \textsc{K-Modes} relies on the \emph{mode}--the most frequently occurring value for each feature. It uses the \textsc{Simple Matching Dissimilarity} measure \citep{dinh2024categorical} to quantify the distance between categorical objects.
The algorithm begins by randomly selecting k data points as the initial centroids. It then assigns each data point to the cluster whose centroid has the minimum dissimilarity, often measured by the number of mismatched categorical attributes. After the assignments, the centroids are updated by determining the mode for each feature within the cluster. This process of assignment and update is repeated iteratively until the centroids stabilize or a predefined stopping criterion is met. \textsc{K-Modes} is computationally efficient and well-suited for categorical data. Still, it requires pre-specifying the number of clusters $k$ and can be sensitive to the initial choice of centroids.

\textsc{K-Prototypes} \citep{huang1998extensions} is a clustering algorithm designed for mixed-type data, combining the principles of \textsc{K-Means} and \textsc{K-Modes}. It uses an adapted distance function that treats each data type separately, updating numeric variables with their means and categorical variables with their modes. It efficiently handles mixed data in one algorithm, rather than running separate clustering processes for each data type. 
 
\textsc{Fuzzy C-Means (FCM)} \citep{bezdek1984fcm} is a clustering algorithm that assigns data points to multiple clusters with varying degrees of membership, rather than forcing each point into a single cluster. FCM minimizes an objective function by iteratively updating cluster centers and fuzzy membership values. This approach is particularly effective for datasets with overlapping clusters, as it allows for soft boundaries between clusters.

\begin{algorithm}[!htb]
\caption{\textsc{Hierarchical Agglomerative Clustering}}
\DontPrintSemicolon
\SetAlgoLined
\KwIn{$D = \{x_i \mid i = 1, 2, \dots, n\}$, initially each $x_i$ is in its own cluster.}

Compute $M = \{d_{ij}\}$, the $(n \times n)$ matrix of pairwise dissimilarities between the clusters, where $d_{ij} = d(x_i, x_j)$.

\Repeat{only one cluster remains}{
    Find the smallest dissimilarity in $D$, say $d_{IJ}$, between two clusters $I$ and $J$\;
    Merge clusters $I$ and $J$ into a new cluster $IJ$\;
    Compute new distances from cluster $IJ$ to any other cluster $K$:
    \begin{itemize}
        \item \textbf{Single linkage:} 
              $d_{IJ,K} = \min(d_{IK}, d_{JK})$,
        \item \textbf{Complete linkage:} 
              $d_{IJ,K} = \max(d_{IK}, d_{JK})$,
        \item \textbf{Average linkage:} 
              $d_{IJ,K} = \frac{\sum_{i \in I \cup J} \sum_{k \in K} d_{ik}}{N_{IJ} \cdot N_{K}}$.
    \end{itemize}
    Update the dissimilarity matrix $M$ by:
    \begin{enumerate}
        \item Remove rows and columns corresponding to clusters $I$ and $J$,
        \item Add a row and column for the new cluster $IJ$,
        \item Fill in the new distances $d_{IJ,K}$ for the updated matrix.
    \end{enumerate}
    
    Record the merge and its distance
}
\textbf{Output:} Dendrogram representing the merge sequence.
\label{algo:hac}
\end{algorithm}

\textsc{Fuzzy K-Modes} \citep{huang1999fuzzy} is an extension of fuzzy clustering designed for categorical data. While \textsc{Fuzzy C-Means} operates on numerical data and relies on Euclidean distance, \textsc{Fuzzy K-Modes} replaces this with a dissimilarity measure tailored for categorical attributes such as the \textsc{Simple Matching Dissimilarity} measure \citep{dinh2024categorical}. It calculates cluster modes rather than means, while still allowing fuzzy membership, making it ideal for datasets involving categorical features.

\begin{algorithm}[!htb]
\caption{\textsc{Partitioning Around Medoids (PAM)}}
\DontPrintSemicolon
\SetAlgoLined

\KwIn{$D = \{x_1, x_2, \ldots, x_n\}$ is the set of data points, $k$ is the number of medoids.}
\textbf{Build phase:}\\
Greedily select $k$ points from $D$ as the medoids to minimize the cost.\\
Assign each $x_i \in D$ to the closest medoid.

\vspace{0.5em}
\textbf{Swap phase:}\\
\While{the cost of the current configuration decreases}{
  \ForEach{medoid $m$}{
    \ForEach{non-medoid $o$}{
      Compute the cost change $\Delta$ for swapping $m$ and $o$.\;
      \If{$\Delta$ is the best cost change so far}{
        Store $m_{\text{best}} \leftarrow m$ and $o_{\text{best}} \leftarrow o$\;
      }
    }
  }
  \If{a swap $m_{\text{best}}$ and $o_{\text{best}}$ reduces the cost}{
    Perform the swap of $m_{\text{best}}$ and $o_{\text{best}}$.\;
  }
  \Else{
    \textbf{break} \tcp*{Stop if no swap reduces the cost.}
  }
}
\label{alg:PAM}
\end{algorithm}

\textsc{Hierarchical Agglomerative Clustering (HAC)} is a bottom-up clustering approach that begins with each data point as an individual cluster and iteratively merges the closest clusters until a single cluster encompassing all data points is formed. The process is typically visualized using a dendrogram, a tree-like diagram that represents the nested grouping of data points and the distances between clusters. The dendrogram can be cut at a chosen level to define the desired clusters. The merging of clusters is determined by a linkage criterion, which specifies how the distance between clusters is calculated. Common linkage methods include \textsc{Single Linkage} \citep{sibson1973slink}, which uses the shortest distance between any two points in different clusters, often resulting in elongated or chain-like clusters; \textsc{Complete Linkage} \citep{defays1977efficient}, which considers the farthest distance between points, producing more compact and spherical clusters; and \textsc{Average Linkage}, which calculates the average distance between all pairs of points across clusters and offers a balance between the two extremes. \textsc{Ward Linkage} \citep{ward1963hierarchical} minimizes the total within-cluster variance by reducing the sum of squared differences across all clusters. This variance minimizing approach is conceptually similar to the \textsc{K-Means} objective but is implemented through an agglomerative hierarchical strategy. HAC is widely applied in bioinformatics, social network analysis, and text mining to uncover hierarchical relationships in complex datasets.

\begin{algorithm}[!htb]
\caption{Abstract DBSCAN \citep{schubert2017dbscan}}
\label{alg:dbscan}
\DontPrintSemicolon
\SetAlgoLined
    Compute neighbors of each point and identify core points \tcp*{Identify core points} 
    Join neighboring core points into clusters \tcp*{Assign core points}
    \ForAll{non-core point}{
        \If{can add to a neighboring core point}{
            Add to the core point \tcp*{Assign border points}}
        \Else{
            Add to noise \tcp*{Assign noise points}
        }
    }
\label{algo:dbscan}
\end{algorithm}

Density-Based Spatial Clustering of Applications with Noise (\textsc{DBSCAN}) \citep{ester1996density} identifies clusters as contiguous regions of high point density, separated by areas of lower density. Algorithm \ref{algo:dbscan} shows the main steps of DBSCAN. The algorithm requires two parameters, epsilon ($\epsilon$), which defines the radius of the neighborhood around a data point, and $minPts$, the minimum number of points required to form a dense region. Data points are classified into three categories: \emph{core points}, which have at least $minPts$ neighbors within their $\epsilon$ radius; \emph{border points}, which lie within the $\epsilon$ neighborhood of a core point but do not meet the $minPts$ threshold themselves; and \emph{noise points}, which are not part of any cluster. Clusters are formed by linking core points and their reachable neighbors iteratively, allowing the algorithm to discover clusters of arbitrary shapes. \textsc{DBSCAN} does not require a pre-defined number of clusters and is robust to noise and outliers. However, its performance depends on the careful selection of $\epsilon$ and $minPts$, as these parameters influence the sensitivity to cluster density and the detection of meaningful patterns in the data.

The \textsc{Genetic K-Means} algorithm \citep{krishna1999genetic} extends the traditional \textsc{K-Means} by integrating genetic operators to improve clustering performance. \textsc{K-Means} is efficient but often sensitive to initial centroid placement and prone to getting stuck in local minima. \textsc{Genetic K-Means} addresses these limitations by using genetic operations like selection, crossover, and mutation to search for better cluster configurations.
The algorithm begins with an initial population of solutions, where each solution represents a set of centroids. The fitness of each solution is evaluated by minimizing intra-cluster variance (sum of squared distances within clusters). The best solutions are selected for reproduction, and genetic operators generate new solutions, enabling a global search for optimal centroids. Over multiple generations, the algorithm evolves toward a configuration that minimizes clustering errors. This global search capability makes \textsc{Genetic K-Means} more robust, especially for complex datasets with noise, outliers, or irregular cluster shapes.

\textsc{Clustering in QUEst} (CLIQUE) \citep{agrawal1998automatic} is a subspace clustering algorithm designed for high-dimensional data. It partitions the data space into a grid of cells and identifies clusters by analyzing cell density. Regions with densities above a minimum threshold are considered potential clusters. The algorithm operates in two main stages: first, it identifies dense regions by examining the density of grid cells. Second, it selects dense subspaces containing significant data points. Unlike traditional methods, CLIQUE clusters data in subspaces rather than the full space, making it ideal for high-dimensional datasets. CLIQUE can detect clusters of arbitrary shapes without requiring prior knowledge of the number of clusters. Its scalability and efficiency make it well-suited for large, complex datasets. It should be noted that CLIQUE allows overlapping clustering by identifying clusters independently in various subspaces, permitting a data point to belong to multiple clusters. In contrast, projected clustering \citep{aggarwal1999fast} generates disjoint clusters, assigning each data point to a single cluster within its relevant subspace, resulting in simpler, non-overlapping structures.

\textsc{CluStream} \citep{aggarwal2003framework} is a data stream clustering algorithm designed for large, dynamic, and time-varying data. It operates in two phases: first, it incrementally maintains micro-clusters, which compactly capture local patterns in real time as new data points arrive. Second, it periodically performs macro-clustering by aggregating these micro-clusters to reveal global structures and trends. The algorithm dynamically adapts by merging, splitting, and updating clusters, making it suitable for evolving data streams. Its computational efficiency and low memory usage make \textsc{CluStream} effective for large-scale, real-time clustering applications.

\textsc{Cluster Ensembles} \citep{strehl2002cluster} is a pioneering work in ensemble clustering, formally defining the problem and proposing three consensus functions: CSPA (\textsc{Cluster-based Similarity Partitioning Algorithm}), HGPA (\textsc{HyperGraph Partitioning Algorithm}), and MCLA (\textsc{Meta-CLustering Algorithm}). Unlike earlier studies in phylogenetics and data fusion that explored combining multiple clusterings, this work introduced a mutual information-based optimization framework and methods that enable knowledge reuse without requiring access to original features. The approach generates diverse clustering solutions by employing different algorithms or multiple runs of the same algorithm with varying random initializations (e.g., different starting points for k-means). This diversity enhances the stability and improves the quality of the final clustering, particularly in distributed and noisy data scenarios, making it effective for data with complex patterns or uncertain structures.

\textsc{MClust} \citep{fraley1998mclust} is an \textsc{R} library that implements \textsc{Gaussian Mixture Models (GMM)} for probabilistic and flexible clustering. It assumes data arises from a mixture of Gaussian distributions, with each cluster defined by parameters such as a mean vector, covariance matrix, and mixing coefficient. Using the \textsc{Expectation-Maximization (EM)} algorithm, \textsc{MClust} iteratively calculates cluster membership probabilities and updates Gaussian parameters to maximize data likelihood. Its flexibility stems from the ability to capture diverse shapes and densities via covariance matrices, making it suitable for complex datasets. Additionally, \textsc{MClust} automatically selects the optimal number of clusters using the Bayesian Information Criterion (BIC), enhancing its adaptability and robustness.

\textsc{Louvain Clustering} \citep{blondel2008fast} is a graph-based network community detection method that optimizes modularity, a measure of the density of edges within communities compared to those between them. The algorithm operates in two phases. First, it assigns nodes to communities by iteratively maximizing modularity gains through node movement. Next, these communities are aggregated into super-nodes, and the process is repeated hierarchically. This multi-level approach enables \textsc{Louvain Clustering} to efficiently detect communities at different scales, making it highly scalable and effective for analyzing large, complex networks.

\textsc {Uniform Manifold Approximation and Projection (UMAP)} \citep{mcinnes2018umap} is a dimensionality reduction technique that excels at preserving the local and global structure of high-dimensional data in a lower-dimensional space. It operates by constructing a high-dimensional graph of data points and optimizing its low-dimensional embedding to maintain the graph’s topological properties. This capability makes \textsc{UMAP} an effective preprocessing step for clustering, as it enhances the separation of data clusters by revealing inherent patterns and groupings. Its scalability and flexibility enable \textsc{UMAP} to handle large and complex datasets, making it a popular choice for clustering tasks in fields such as genomics, image analysis, and natural language processing.
Recently, \cite{sun2024riccinet} proposed \textsc{RicciNet}, a deep clustering model based on a Riemannian Generative Model. Like UMAP, this approach leverages Riemannian space instead of the traditional Euclidean space, demonstrating improvements in clustering performance.
\subsection{Libraries, Tools, and Frameworks}
\begin{figure*}[!htb]
\vspace{-2cm}
  \centering
  \includegraphics[width=0.8\linewidth]{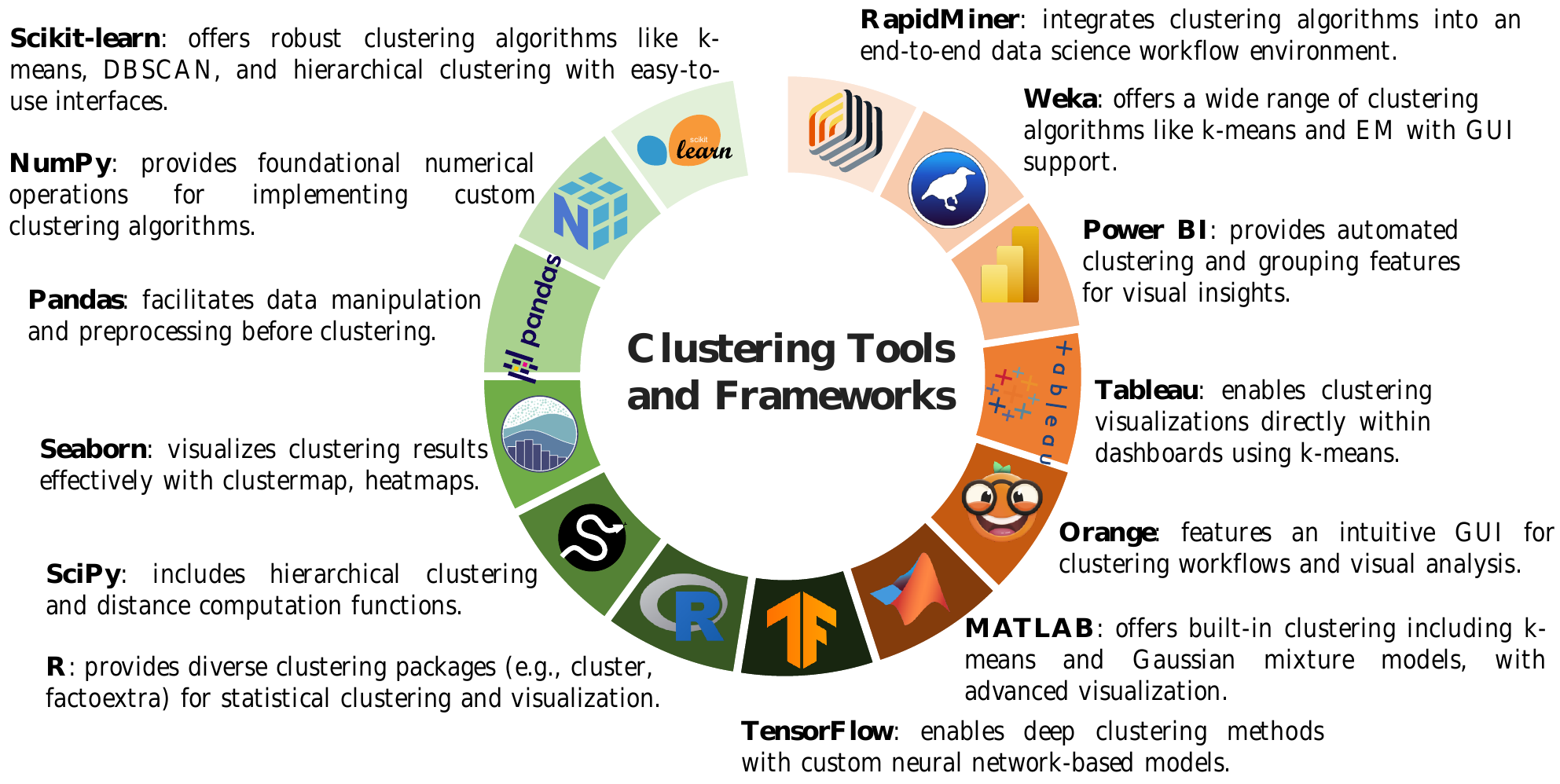}
  \caption{Common tools, libraries and framework for clustering}
  \label{fig:tools}
\end{figure*}

\textsc{Python} has become the most used language on GitHub\footnote{\url{https://github.blog/news-insights/octoverse/octoverse-2024/\#the-most-popular-programming-languages}}, overtaking JavaScript after a 10-year run as the most used language. In \textsc{Python}, clustering is effortless with \textsc{Scikit-learn}\footnote{\url{https://scikit-learn.org/1.5/modules/clustering.html}}, a versatile and user-friendly machine learning library. It provides efficient implementations of popular clustering algorithms like \textsc{K-Means}, \textsc{DBSCAN}, \textsc{HDBSCAN}, \textsc{Hierarchical Clustering}, \textsc{Mean Shift}, and \textsc{Spectral Clustering}, making it a go-to choice for data scientists and engineers. With its intuitive API and seamless integration with libraries such as \textsc{NumPy}\footnote{\url{https://numpy.org/}} and \textsc{Pandas}\footnote{\url{https://pandas.pydata.org/}}, \textsc{Scikit-learn} is a good choice for prototyping and experimenting with different clustering tasks. For instance, performing \textsc{K-Means} clustering can be achieved with just a few lines of code, typically involving steps like data preprocessing, model fitting, and result visualization.

\textsc{R}\footnote{\url{https://www.r-project.org/}} is a statistical programming language renowned for its strength in exploratory data analysis and visualization. Libraries like \emph{cluster}\footnote{\url{https://cran.r-project.org/web/packages/cluster/index.html}} and \emph{factoextra}\footnote{\url{https://cran.r-project.org/web/packages/factoextra/index.html}} provide robust tools for implementing clustering techniques, including \textsc{K-Means}, with an emphasis on visually interpreting cluster distributions. Its capabilities make \textsc{R} a preferred choice for academic research and projects requiring statistical rigor and insightful presentation.

\begin{figure*}[!htb]
\vspace{-2cm}
  \centering
  \includegraphics[width=0.8\linewidth]{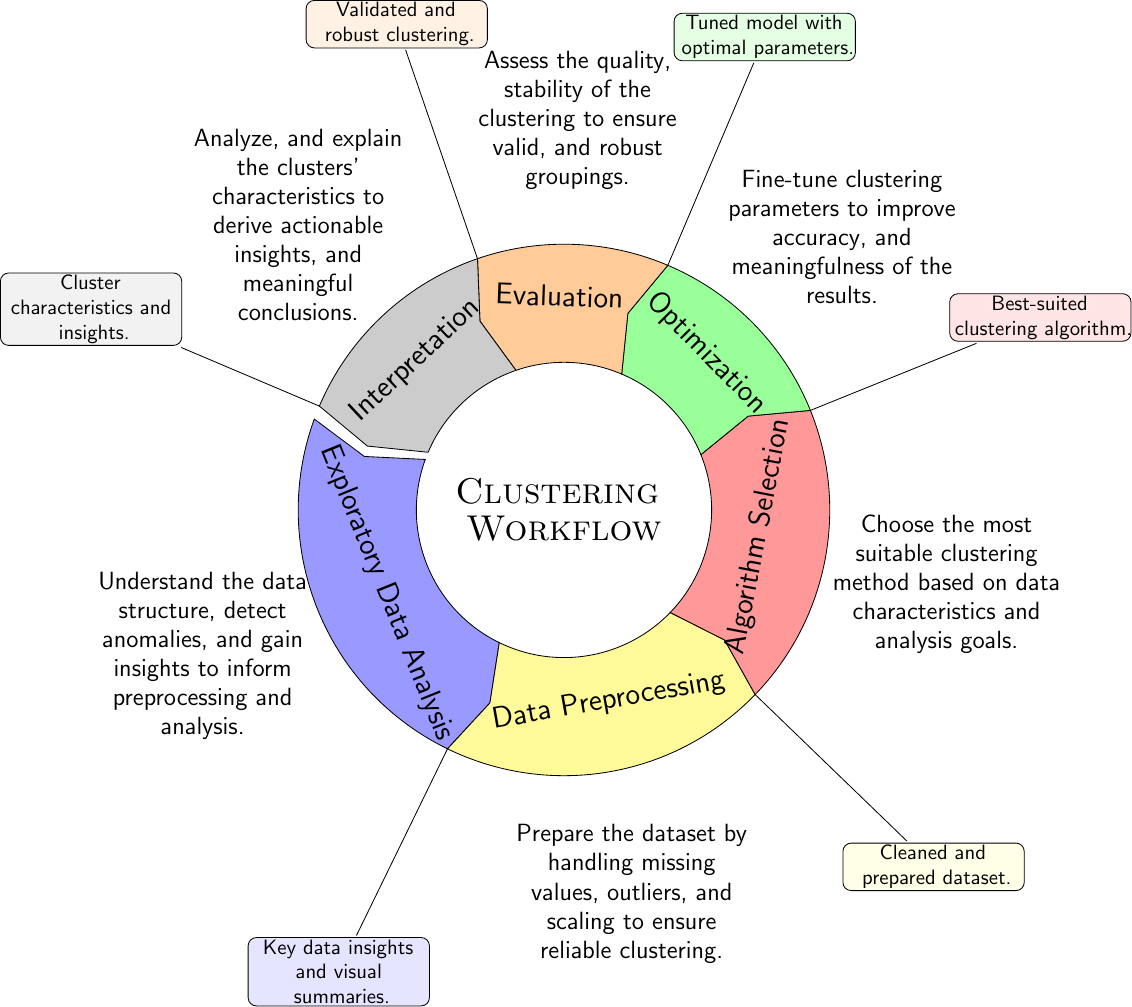}
  \caption{A general workflow of clustering tasks}
  \label{fig:workflow}
\end{figure*}

\begin{figure*}[!htb]
  \centering
  \includegraphics[width=\linewidth]{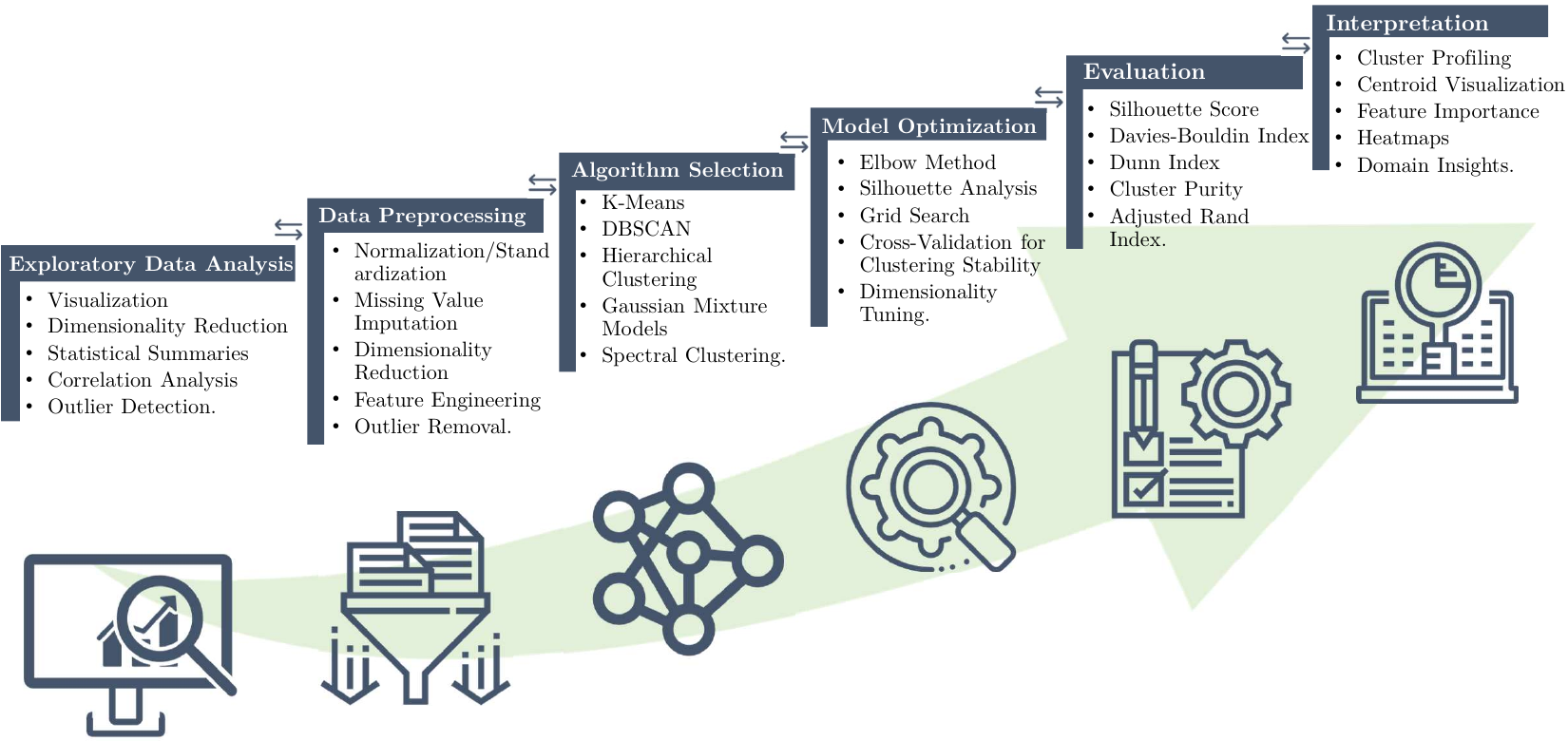}
  \caption{A step-by-step action in the workflow of clustering}
  \label{fig:waterfall}
\end{figure*}

\textsc{MATLAB} is a high-level language designed for numerical computation and visualization, offering built-in functions for clustering\footnote{\url{https://www.mathworks.com/help/stats/cluster-analysis.html?s_tid=CRUX_lftnav}} such as \textsc{K-Means}, \textsc{Hierarchical Clustering}, \textsc{K-Medoids}, \textsc{DBSCAN}, \textsc{Spectral Clustering}, \textsc{Gaussian Mixture Models} and \textsc{Affinity Propagation Clustering}. Widely used in academia and industries like engineering and finance, \textsc{MATLAB} enables precise modeling and insightful visualizations. For instance, its hierarchical clustering tools allow users to visualize cluster relationships with dendrograms, providing an intuitive way to explore data structure.

\textsc{Orange Data Mining} \footnote{\url{https://orangedatamining.com/}} is an open-source platform for data visualization and analysis, offering widges, drag-and-drop interface for tasks like preprocessing, clustering, and classification. It supports clustering methods such as \textsc{K-Means}, \textsc{Hierarchical Clustering}, \textsc{DBSCAN}, and \textsc{Network Clustering}, with tools for parameter tuning, cluster visualization, and quality assessment. Advanced visualizations like dendrograms and scatter plots make it easy to explore and interpret cluster assignments. Its integration with \textsc{Python} scripting and add-ons further extends its versatility for machine learning and statistical analysis.

\textsc{Tableau}\footnote{\url{https://help.tableau.com/current/pro/desktop/en-us/clustering.htm\#how-clustering-works}} leverages clustering to identify patterns and group similar data points within visualizations. With its built-in \textsc{K-Means} clustering, users can segment data directly in dashboards, enabling deeper insights into customer behavior, market trends, or performance metrics without requiring extensive coding.

\textsc{Power BI}\footnote{\url{https://learn.microsoft.com/en-us/analysis-services/data-mining/microsoft-clustering-algorithm?view=asallproducts-allversions}} includes clustering as part of its data analytics features, using algorithms like \textsc{K-Means} and \textsc{Expectation Maximization} to uncover hidden patterns in datasets. Clustering helps users segment data dynamically within reports and dashboards, making it ideal for tasks like customer segmentation, sales analysis, and anomaly detection.

\textsc{Weka}\footnote{\url{https://www.tutorialspoint.com/weka/weka_clustering.htm}} provides an accessible platform for clustering through its extensive machine learning library, supporting methods like \textsc{K-Means}, \textsc{DBSCAN}, \textsc{Hierarchical Clustering} and \textsc{EM}. Users can apply clustering to explore and analyze data patterns, often as a preprocessing step for further classification or prediction tasks.

\textsc{RapidMiner}\footnote{\url{https://docs.rapidminer.com/latest/studio/operators/index.html}} offers clustering tools for tasks like market segmentation and outlier detection. With support for various techniques such as \textsc{K-Means} and \textsc{Hierarchical Clustering}, \textsc{DBSCAN}, and \textsc{K-Medoids}, users can combine clustering workflows with advanced machine learning models for end-to-end data analysis.

\subsection{Clustering workflow in data science}
Figure \ref{fig:workflow} illustrates the overall workflow of the clustering process, while Figure \ref{fig:waterfall} highlights various options available for each step in the workflow. The first step, \textsc{Exploratory Data Analysis (EDA)}, involves examining the dataset to understand its structure and distribution. During this phase, patterns, anomalies, and outliers are detected, and key insights are derived using visualization and statistical techniques. These insights provide valuable context and help inform the subsequent steps of the process.

Following EDA, \textsc{Data Preprocessing} ensures the dataset is prepared for clustering by addressing issues such as missing values, outliers, and differing scales. Techniques like standardization or normalization, for instance, can help to align feature scales, while handling inconsistencies improves data quality. However, the specific preprocessing steps required depend on the nature of the dataset and the clustering algorithm being used. Proper preprocessing is essential to minimize the risk of biased or unreliable results.

The third step, \textsc{Algorithm Selection}, focuses on identifying the most appropriate clustering method based on the dataset characteristics and the analysis goals. Depending on data complexity, cluster shapes/characteristics, and computational requirements, algorithms like \textsc{K-Means}, DBSCAN, or hierarchical clustering may be chosen. The choice of algorithm lays the foundation for producing meaningful groupings.

After selecting an algorithm, \textsc{Optimization} is performed to fine-tune parameters, such as the number of clusters or distance metrics. Adjustments at this stage enhance the clustering's accuracy and ensure that the results are both precise and interpretable. Fine-tuning the model makes the clusters more relevant to the underlying data structure.

In the \textsc{Evaluation} phase, the quality and stability of the clusters are assessed using metrics such as silhouette score, Dunn index, or cohesion and separation measures. This ensures that the groupings are valid, robust, and aligned with the intended analysis goals. Evaluation is critical for verifying the clustering's reliability and significance.

Finally, the \textsc{Interpretation} step involves analyzing the characteristics of each cluster to derive actionable insights. By understanding the unique attributes and patterns within each group, meaningful conclusions can be drawn, aiding decision-making or further research. This step transforms raw data clusters into comprehensible and valuable insights.

Altogether, this workflow ensures a systematic approach to clustering, moving from raw data exploration to actionable insights with clarity and rigor.
\section{Challenges in practical implementation} \label{sec:challenges}

\begin{figure*}[!htb]
\vspace{-2.5cm}
  \centering
  \includegraphics[width=\linewidth]{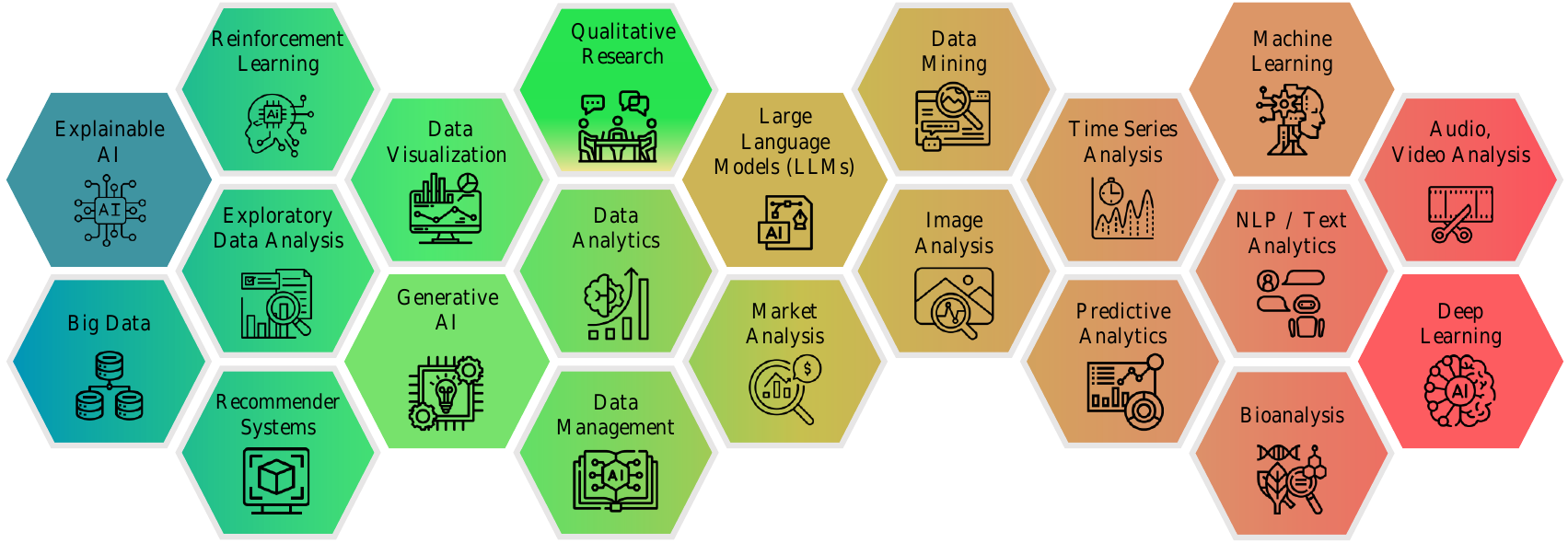}
  \caption{Applications of data clustering in different topics}
  \label{fig:apps_clustering}
\end{figure*}

\subsection{Data-related challenges}
Clustering high-dimensional data is challenging due to the \emph{curse of dimensionality}, where distances lose interpretability, making distance-based methods less effective \citep{peng2023interpreting}. Dimensionality reduction techniques such as Principal Component Analysis (PCA), t-distributed Stochastic Neighbor Embedding (t-SNE), and UMAP help capture key features, while alternative methods like \textsc{Subspace Clustering}, such as CLIQUE, perform better in high dimensions.

Clustering large-scale datasets is computationally intensive. Algorithms like \textsc{Hierarchical Clustering} scale poorly, while \textsc{Mini-Batch K-Means}\footnote{\url{https://scikit-learn.org/stable/modules/generated/sklearn.cluster.MiniBatchKMeans.html}}, which processes small random subsets, and distributed frameworks like \textsc{Apache Spark} address scalability by enabling parallel computing. Preprocessing steps, such as dimensionality reduction, further helps managing data size.

Noise and outliers distort cluster assignments, particularly in \textsc{K-Means}, where centroids are shifted \citep{iam2020clustering}. Algorithms like \textsc{DBSCAN} handle noise better by marking outliers, though their performance depends heavily on parameter tuning, especially with varying cluster densities. Preprocessing with outlier detection techniques, such as Isolation Forest\footnote{\url{https://scikit-learn.org/1.5/modules/generated/sklearn.ensemble.IsolationForest.html}}, helps remove outliers before clustering.

Missing data disrupts clustering by affecting distance calculations. Multiple imputation techniques, which account for uncertainty by generating multiple plausible datasets, are generally preferred for addressing this issue. Simpler imputation methods, such as replacing missing values with the mean or median, may be used in less critical scenarios but risk introducing bias. In cases of excessive missing data, removing features or records might be unavoidable, or clustering methods that can handle missing data directly can be applied \citep{dinh2021clustering}.
\subsection{Algorithms-related challenges}
Algorithm selection is crucial, as different clustering methods excel under specific conditions. For instance, \textsc{K-Means} is effective for spherical clusters but struggles with varying densities or irregular shapes, while \textsc{DBSCAN} handles arbitrary shapes and noise but depends heavily on parameter tuning. \textsc{Hierarchical Clustering} offers detailed, dendrogram-based insights but is computationally expensive for large datasets. Visualization techniques, such as t-SNE or UMAP, can aid in exploring the data structure. Comparing algorithms using metrics like the \textsc{Silhouette} score or \textsc{Davies-Bouldin} index can guide the selection of the most suitable method for achieving meaningful clustering results tailored to the analysis goals.

Algorithm settings also play a vital role, as clustering outcomes are sensitive to hyperparameters. \textsc{K-Means} requires specifying the number of clusters ($k$), \textsc{DBSCAN} relies on $\epsilon$ (distance threshold) and $minPts$ (minimum points per cluster), and \textsc{Hierarchical Clustering} depends on linkage criteria. Optimal tuning of these settings often involves methods like \textsc{Elbow} plots, \textsc{Silhouette} analysis \citep{dinh2019estimating}, or advanced techniques such as grid search and Bayesian optimization, supported by domain expertise for improved results.

\section {Applications of data clustering in data science} \label{sec:applications}
Clustering plays an important role across diverse domains in data science, enabling insightful analysis and efficient solutions. Figure \ref{fig:apps_clustering} illustrates potential applications of clustering in data science, which are discussed in detail below.

In \textsc{Explainable AI}, clustering reveals underlying patterns in data, making complex models more interpretable and enhancing transparency in decision-making processes \citep{alvarez2024comprehensive}. In \textsc{Big Data Analysis}, clustering organizes massive and unstructured datasets into manageable groups, facilitating anomaly detection, customer segmentation, and pattern discovery, which are essential for industries like finance, healthcare, and social media \citep{nasraoui2019clustering}. \textsc{Recommender Systems} leverage clustering to group users or items based on behavioral similarities, enabling personalized recommendations that improve user engagement and system performance \citep{gasparetti2021community}. Similarly, in \textsc{Reinforcement Learning}, clustering simplifies complex environments by grouping states or actions with similar outcomes, reducing dimensionality and expediting optimal policy learning \citep{chang2024improved}.

\begin{figure}[!htb]
  \begin{subfigure}[!htb]{0.49\columnwidth}
    \includegraphics[width=\columnwidth]{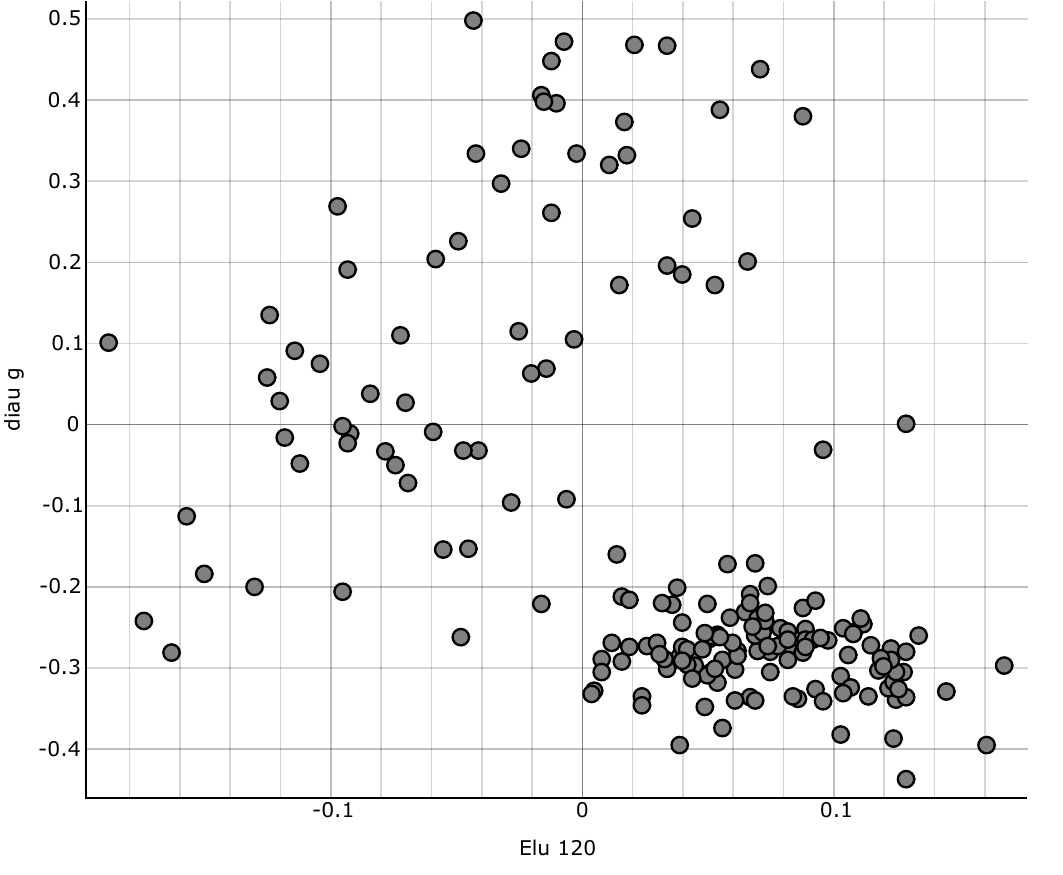}
    \caption{Scatter plot of the original data}
    \label{fig:1-scatter_origin}
  \end{subfigure}
  \begin{subfigure}[!htb]{0.49\columnwidth}
    \includegraphics[width=\columnwidth]{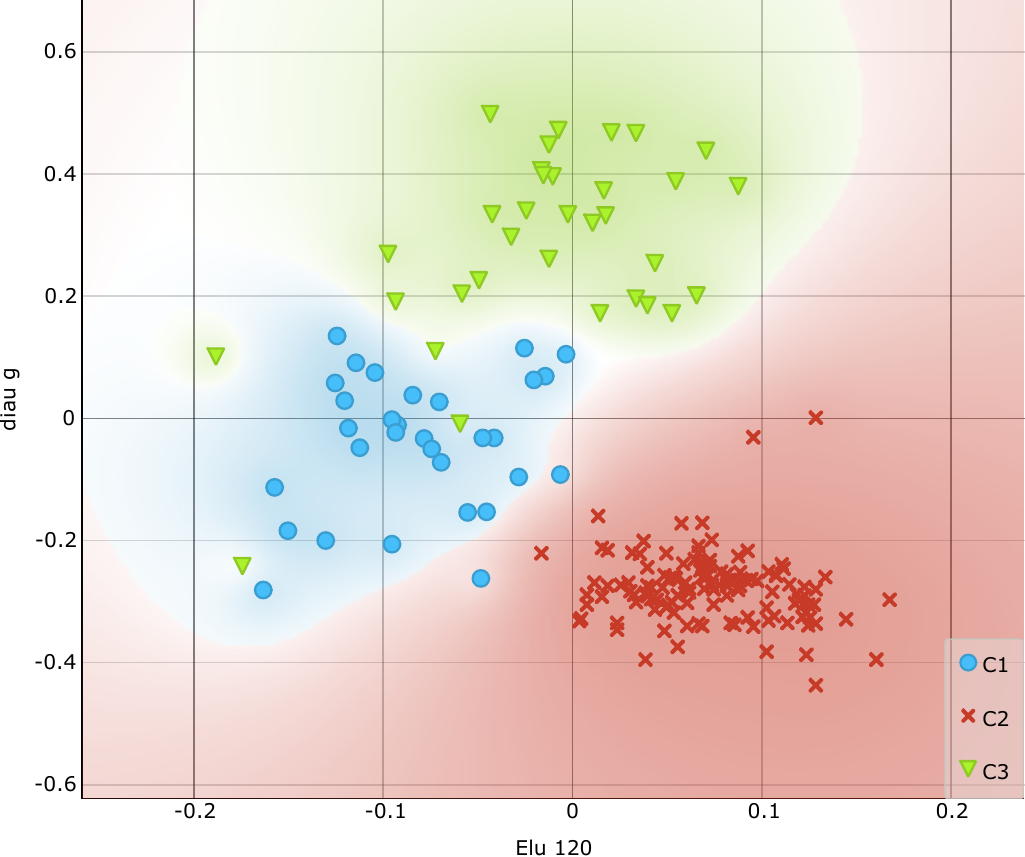}
    \caption{Scatter plot of clustering results}
     \label{fig:1-scatter_clustering}
  \end{subfigure}
  \caption{\textsc{K-Means} clustering for the Brown data set}
  \label{fig:1-clutering_example}
\end{figure}

\begin{figure*}[!htb]
\vspace{-2.5cm}
  \centering
  \includegraphics[width=0.9\linewidth]{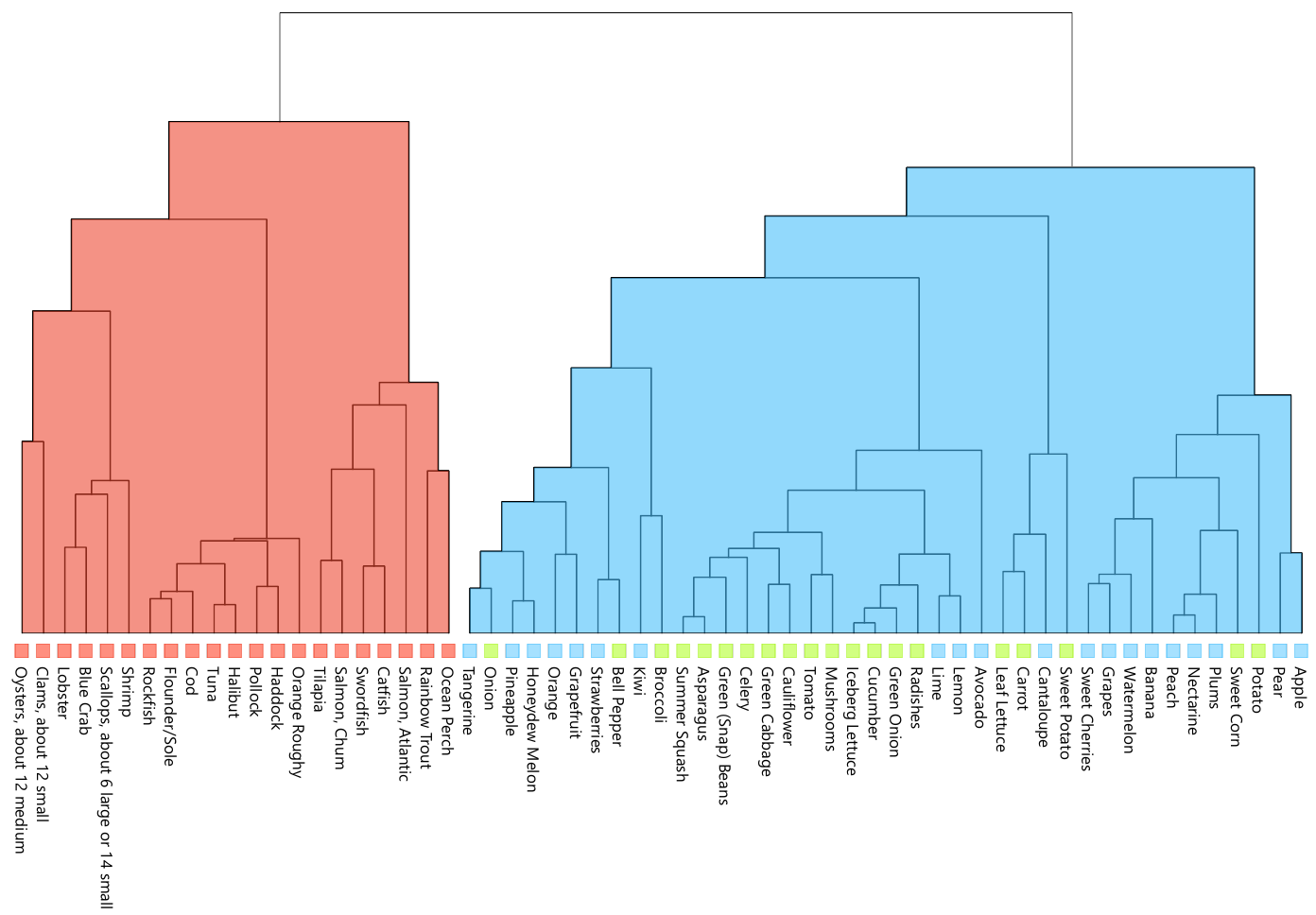}
  \begin{tabular}{ccc}
    \squaremarker{SeafoodColor} Seafood &
    \hspace{1cm} \squaremarker{FruitsColor} Fruits &
    \hspace{1cm} \squaremarker{VegetablesColor} Vegetables \\
  \end{tabular}
  \caption{Hierarchical clustering for Food Nutrition Information dataset}
  \label{fig:hac_clustering}
\end{figure*}

Clustering is integral to \textsc{Exploratory Data Analysis (EDA)}, as it identifies hidden structures and relationships, such as customer segments or patient subgroups, providing a foundation for further analysis \citep{tukey1977exploratory}.
For \textsc{Data Visualization}, clustering organizes high-dimensional data into interpretable groups, aiding techniques like t-SNE and UMAP to highlight patterns and simplify analysis. Tools such as dendrograms, scatterplots, and heatmaps often accompany clustering methods to provide intuitive representations, allowing users to gain insights and guide further analysis or decision-making \citep{waskom2021seaborn}. Figure \ref{fig:1-clutering_example} illustrates how \textsc{K-Means} performs clustering to segment a biological dataset named Brown dataset\footnote{\url{https://datasets.biolab.si/core/brown-selected.tab}} into three clusters. 
Figure \ref{fig:hac_clustering} illustrates the use of \textsc{Hierarchical Clustering} with Ward's linkage and normalized Euclidean distance to organize the nutrition food dataset\footnote{\url{https://www.kaggle.com/datasets/cid007/food-and-vegetable-nutrition-dataset}} into a hierarchical structure.

In \textsc{Generative AI}, clustering enhances efficiency by segmenting datasets into meaningful subsets, improving model outputs such as realistic image generation or coherent text generation. Clustering also optimizes \textsc{Data Management} by grouping similar records for better retrieval, deduplication, and quality assurance, especially in large-scale databases \citep{garcia2017clustering}.

In \textsc{Data Analytics}, clustering enables actionable insights by segmenting datasets into homogeneous groups, aiding in targeted marketing, customer retention strategies, and behavioral analysis \citep{reutterer2021cluster}. \textsc{Large Language Models (LLM)} benefit from clustering, which organizes text data into semantically similar groups, aiding in model fine-tuning, contextual understanding, summarization, and mitigating hallucinations in LLMs \citep{he2024mitigating}. For \textsc{Market Analysis}, clustering identifies distinct customer segments and market trends, empowering businesses to design personalized campaigns, optimize pricing strategies, and understand competitor dynamics \citep{muller2014stability}.

Clustering is a core technique in \textsc{Data Mining}, uncovering hidden patterns in data for anomaly detection, trend analysis, and knowledge discovery across diverse fields \citep{berkhin2006survey}. In \textsc{Image Analysis}, clustering groups similar pixels or features, enabling tasks like image segmentation, object recognition, and medical diagnostics. Recent advancements in LLMs have transformed image-based clustering by enabling feature extraction through embeddings. These embeddings represent high-dimensional image features in a compact vector format, capturing semantic relationships effectively \citep{huang2024deep}. Once features are extracted, traditional clustering techniques like \textsc{K-Means} or \textsc{DBSCAN} are applied in the embedding space to group similar images \citep{khan2025enhanced}. \textsc{Time Series Analysis} employs clustering to group similar temporal patterns, aiding in forecasting, anomaly detection, and understanding trends in domains like finance, retail, energy, and engineering \citep{aghabozorgi2015time,inoue2024clustering}. \textsc{Predictive Analytics} uses clustering to create meaningful data groups, enhancing targeted predictions in areas like healthcare, where it supports patient outcome predictions \citep{ganapathy2024brain}, and in customer analytics for churn prediction \citep{rajamohamed2018improved}.

In \textsc{Natural Language Processing (NLP)}, clustering identifies themes, topics, and semantic relationships by grouping similar words, sentences, or documents, thereby streamlining tasks like sentiment analysis and topic modeling \citep{cozzolino2022document}. For sentiment analysis, clustering groups text data, such as customer reviews or social media posts, based on their emotional tone or underlying themes \citep{yu2023mining,yu2023understanding}. Similarly, in social network analysis, clustering is instrumental in community detection, uncovering subgroups of interconnected individuals or entities within a larger network \citep{malliaros2013clustering}. These applications provide valuable insights into behavioral trends, opinion mining, and the dynamics of social ecosystems.

\textsc{Bioanalysis} relies heavily on clustering to group genes, classify cell types, and detect protein families, advancing research in personalized medicine and drug discovery \citep{udrescu2016clustering}. In personalized medicine, clustering is used to group patients based on genetic markers, symptoms, or treatment responses, aiding in the design of tailored therapeutic strategies. By identifying distinct biological/clinical patterns, clustering facilitates a deeper understanding of complex biological systems and supports precision healthcare initiatives \citep{lisik2025artificial}.

Clustering also supports \textsc{Machine Learning} by reducing dimensionality, initializing models, and identifying outliers, which enhances model robustness and performance \citep{yang2021mean}. For \textsc{Audio and Video Analysis}, clustering separates similar signals or frames, enabling speech recognition, speaker diarization, activity segmentation, and anomaly detection in videos \citep{qiu2024video}. Clustering complements \textsc{Deep Learning} workflows by organizing data for efficient training or unsupervised representation learning. By clustering neural network embeddings, it uncovers meaningful patterns, validates models, and identifies latent structures in complex datasets \citep{yang2017towards}.

Finally, clustering is a valuable tool in \textsc{Qualitative Research}, as it helps identify patterns and groupings within complex, unstructured data \citep{macia2015using}. Specifically, clustering can be used to group similar themes, narratives, or codes from interviews, focus groups, or textual data, enabling researchers to uncover shared experiences or perspectives. In \textsc{Mixed Quantitative and Qualitative Research}, clustering bridges the two approaches by integrating quantitative data, such as survey responses, with qualitative insights, like open-ended questions, to group participants or phenomena based on combined attributes \citep{peladeau2021cluster}. This enhances the depth of analysis, enabling researchers to explore nuanced patterns and relationships while maintaining a holistic view of the dataset \citep{henry2015clustering}.
\section{Conclusion} \label{sec:conclusion}
Data clustering is an important and commonly used technique in data science, serving as a cornerstone for organizing complex datasets into coherent groups based on intrinsic similarities. As demonstrated in this paper, clustering not only simplifies high-dimensional data but also provides a robust framework for identifying latent structures and patterns that are critical for exploratory analysis and downstream tasks. Its flexibility and applicability across a broad spectrum of data types and domains underscore its enduring significance in solving intricate data challenges.

The integration of clustering with emerging technologies, particularly LLMs represents a significant advancement in the field. LLMs, with their ability to generate high-quality embeddings and capture nuanced patterns in data, enhance clustering by providing more meaningful and compact feature representations. These embeddings are particularly useful for unstructured data, such as text and images, enabling traditional clustering algorithms, such as \textsc{K-Means} to operate effectively in reduced-dimensional spaces. This synergy between clustering and LLMs opens new avenues for applications ranging from image-based analysis to text mining, where extracting semantic relationships and organizing information is essential.

\section*{Declaration of competing interest}
The authors declare that they have no known competing financial interests or personal relationships that could have appeared to influence the work reported in this paper.

\bibliographystyle{model5-names}
\biboptions{authoryear}
\bibliography{_main}
\end{document}